\pdfoutput=1
\documentclass{article}

\usepackage{arxiv}

\usepackage[utf8]{inputenc} % allow utf-8 input
\usepackage[T1]{fontenc}    % use 8-bit T1 fonts
\usepackage{hyperref}       % hyperlinks
\usepackage{url}            % simple URL typesetting
\usepackage{nicefrac}       % compact symbols for 1/2, etc.
\usepackage{microtype}      % microtypography
\usepackage{lipsum}

\usepackage{amsmath,amssymb,amsfonts} % higher quality rendering of eqs
\usepackage{flushend}
\usepackage{booktabs} % higher quality tables
\usepackage{algorithm} % provides algorithm env.
\usepackage{algorithmicx}
\usepackage{algpseudocode}
\usepackage{tikz}
\usepackage{setspace}
\usepackage{array}
\usepackage{multicol}
\usepackage{graphicx}
\usepackage{subfigure}
\usepackage{enumitem}
\usepackage{tabularx}

\title{Detection and Prediction of Cardiac Anomalies Using Wireless Body Sensors and Bayesian Belief Networks}

\author{
  Asim Darwaish\thanks{Use footnote for providing further
    information about author (webpage, alternative
    address)---\emph{not} for acknowledging funding agencies.} \\
  Department of Computer Science\\
  Paris Descartes University\\
  PARIS, 75006 \\
  \texttt{asim.darwaish@etu.parisdescartes.fr} \\
   \And
 Farid Na\"it-Abdesselam \\
  Department of Computer Science\\
  Paris Descartes University\\
  PARIS, 75006 \\
  \texttt{naf@iastate.edu} \\
  \And
 Ashfaq Khokhar \\
  Department of Electrical and Computer Engineering\\
   Iowa State University\\
  Ames, IA 50011 \\
  \texttt{ashfaq@iastate.edu} \\
}

\begin{document}
\maketitle

\begin{abstract}
Intricating cardiac complexities are the primary factor associated with healthcare costs and the highest cause of death rate in the world. However,  preventive measures like the early detection of cardiac anomalies can prevent severe cardiovascular arrests of varying complexities and can impose a substantial impact on healthcare cost.
Encountering such scenarios usually the electrocardiogram (ECG or EKG) is the first diagnostic choice of a medical practitioner or clinical staff to measure the electrical and muscular fitness of an individual heart. 
However, the advancement and the foster development of wearable devices and ECG  sensors make it possible to continuously measure and analyze the electrocardiogram and myocardium signals of the heart. These advancements are also making the early detection of cardiac anomalies useful and alert the respective individual and clinicians to provide better healthcare services. 
The ECG test is the more straightforward test to perform and generally, the clinicians' record this test after encountering the symptoms related to the heart. However, this test requires significant training and adroitness to interpret the recorded ECG correctly and effectively. 
This paper presents a system which is capable of reading the recorded ECG and predict the cardiac anomalies without the intervention of a human expert. The paper purpose an algorithm which read and perform analysis on electrocardiogram datasets.
The proposed architecture uses the Discrete Wavelet Transform (DWT) at first place to perform preprocessing of ECG data followed by undecimated Wavelet transform (UWT) to extract nine relevant features which are of high interest to a cardiologist. The probabilistic mode named Bayesian Network Classifier is trained using the extracted nine parameters on UCL arrhythmia dataset. The proposed system classifies a recorded heartbeat into four classes using Bayesian Network classifier and Tukey's box analysis. The four classes for the prediction of a heartbeat are (a) Normal Beat, (b) Premature Ventricular Contraction (PVC) (c) Premature Atrial Contraction (PAC) and (d) Myocardial Infarction. The results of experimental setup depict that the proposed system has achieved an average accuracy of 96.6 for PAC\% 92.8\% for MI and 87\% for PVC, with an average error rate of 3.3\% for PAC, 6\% for MI and 12.5\% for PVC on real electrocardiogram datasets including Physionet and European ST-T  Database  (EDB).
\end{abstract}

% keywords can be removed

\section{Introduction}\label{sec:introduction}
According to the report generated by World Health Organization~\cite{who}, Cardiac arrest or cardiovascular diseases are the perceptible reason for mortality rate all over the world.  According to the statistics of this report, 17.3 million died in 2008, and 15.2 million died in 2016 due to Ischaemic heart disease and stroke. However, the predicted number for 2030 is 23.3 million which is an alarming indicator for medical care. Cardiovascular diseases are the group of syndromes mainly concerning to heart and blood vessels and sometimes the addition of atherosclerosis results in a stroke. Early detection of cardiac anomalies and relevant preventive measures can significantly decrease the mortality rate caused by CVDs and associated healthcare cost.   It is imperative to mention that out of overall mortality rate linked with CVD, 47\% of death happens out of cardiac care units. Out of cardiac care, monitoring triggered a massive interest in the development of wearable devices and sensors for the proactive healthcare monitoring system.\par
\begin{figure}[!htb]
\centering
\includegraphics[scale=0.60]{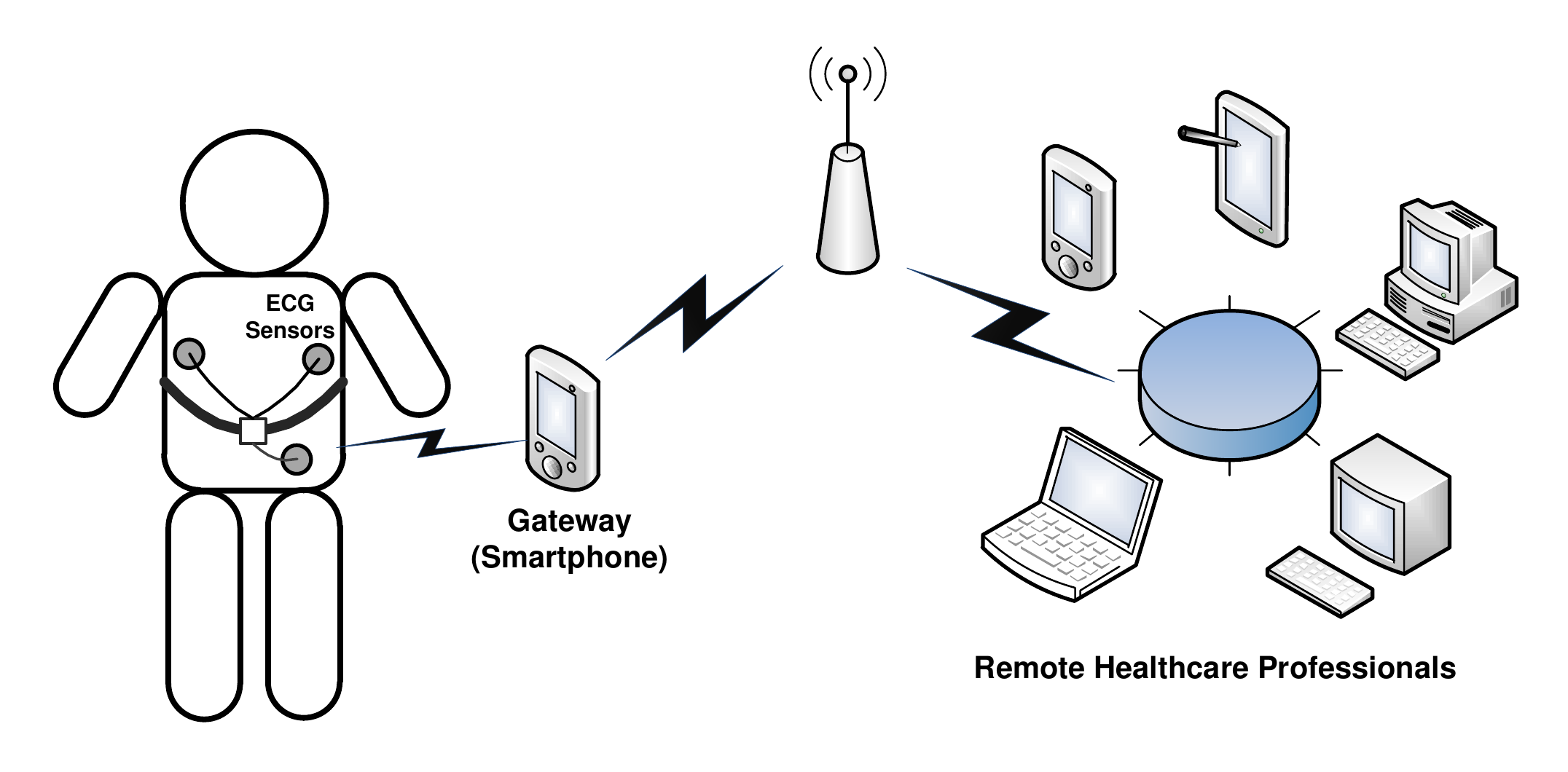}
\caption{A WBAN architecture for ECG monitoring} 
\label{fig:wban}
\end{figure}

 
Whenever a person encounters any cardiac anomaly or a symptom pertaining to CVD, a cardiologist or practitioner usually checks his/her physiological parameters like blood pressure, oxygen saturation, heart rate, and analyze electrocardiogram. Even the patients admitted under cardiac care unit after having cardiac arrest are being monitored continuously for aforesaid physiological signals. However, clinicians required numerous devices like ECG, Holter, pulse oximeter and many other devices for the observation of these physiological parameters after encountering the CVD.  Thanks to the latest industrial advancement and research community for the manufacturing of biomedical sensors (conception and miniaturization) make these devices capable of the collection and transmission of health data remotely using wireless communication. These devices are common for recording numerous physiological signals for instance Oxygen Saturation, Temperature, Heart Rate, Blood Pressure,  ElectroMyoGram, ElectroCardioGram, etc. 
In the traditional health care monitoring system these wireless sensors are employed in WBAN (Wireless Body Area Network). These sensors are affixed to specific human body parts a for the collection of the relevant physiological parameter, wearable comfort and convenience. These sensors transmit the collected data to a cloud or other computational resources or healthcare professionals via gateway device (e.g., smartphone, tablet, .etc.) for the sake of further insights,  expert opinions, analysis, and automated anomaly detection as depicted in Figure \ref{fig:wban}
\par  
 Although the recent technical advancement in the integrated circuits and WBANs are escalating the quality of life, remote monitoring, real-time health care services across numerous health care applications. Similarly, these body sensors and wearable devices are widely used for CVD monitoring. However, these devices are not at their efficacy and need specific challenges. One of the prominent challenges in respect of CVD is automatic and early detection of a cardiac anomaly. Early detection and prediction of associated cardiac anomaly can save a life and reduce healthcare cost substantially. 
There are various kinds of CVDs, for instance, Arrhythmia is a CVD that is related to irregular heartbeats. For a healthy person, the heartbeat is from 60 to 100, but sometimes heart starts beating faster or at a languid pace which is called Arrhythmia or dysrhythmia. Arrhythmia is further classified into tachycardia (very fast heartbeat) and bradycardia (very slow heartbeat). Arrhythmia impacts the rhythm in various forms commonly PAC and PVC. The PAC and PVC are triggered due to premature expulsion of electrical impulses in the atrium ~\cite{PVC}~\cite{PAC}. Myocardial Infarction (MI) is another severe heart anomaly also known as heart failure.  MI mostly triggered due to an inadequate supply of blood and oxygen to heart muscles. The main reason for this improper transportation of blood and oxygen is narrowed and blocked arteries. MI is an acutely ischemic heart disease mostly damages a heart portion which is termed as necrosis. In order to get more details about CVDs disease, one can go deep down into the paper ~\cite{morris2009abc}. \par  
For the verification of a CVD symptom, a cardiologist or a clinician advise for a series of examinations. Usually, the  ECG is the first diagnostic tool to check irregular rhythms. Generally, ECG is transcribed on a graph which represents the contraction and relaxation of cardiac muscles due to depolarization and repolarization of myocardial cells.  These electrical changes are recorded via electrodes placed on the limbs and chest wall  of a patient.
The most commonly used ECG technique is 12- lead ECG ~\cite{morris2009abc}. These leads are placed on 10 sites of human body for the recording of 12 ECG signals. The 12-leads ECG is described as : 
\begin{itemize}[leftmargin=*,label={--}]
\item V1, V2, V3, V4, V5, V6 collectively called as Precordial Leads.
\item I, II, III known as Limb Leads.
\item aVR, aVL, aVF recognized as Limb Leads.
\end{itemize}
\par
\begin{figure}[!htb]
\centering
\includegraphics[scale=0.60]{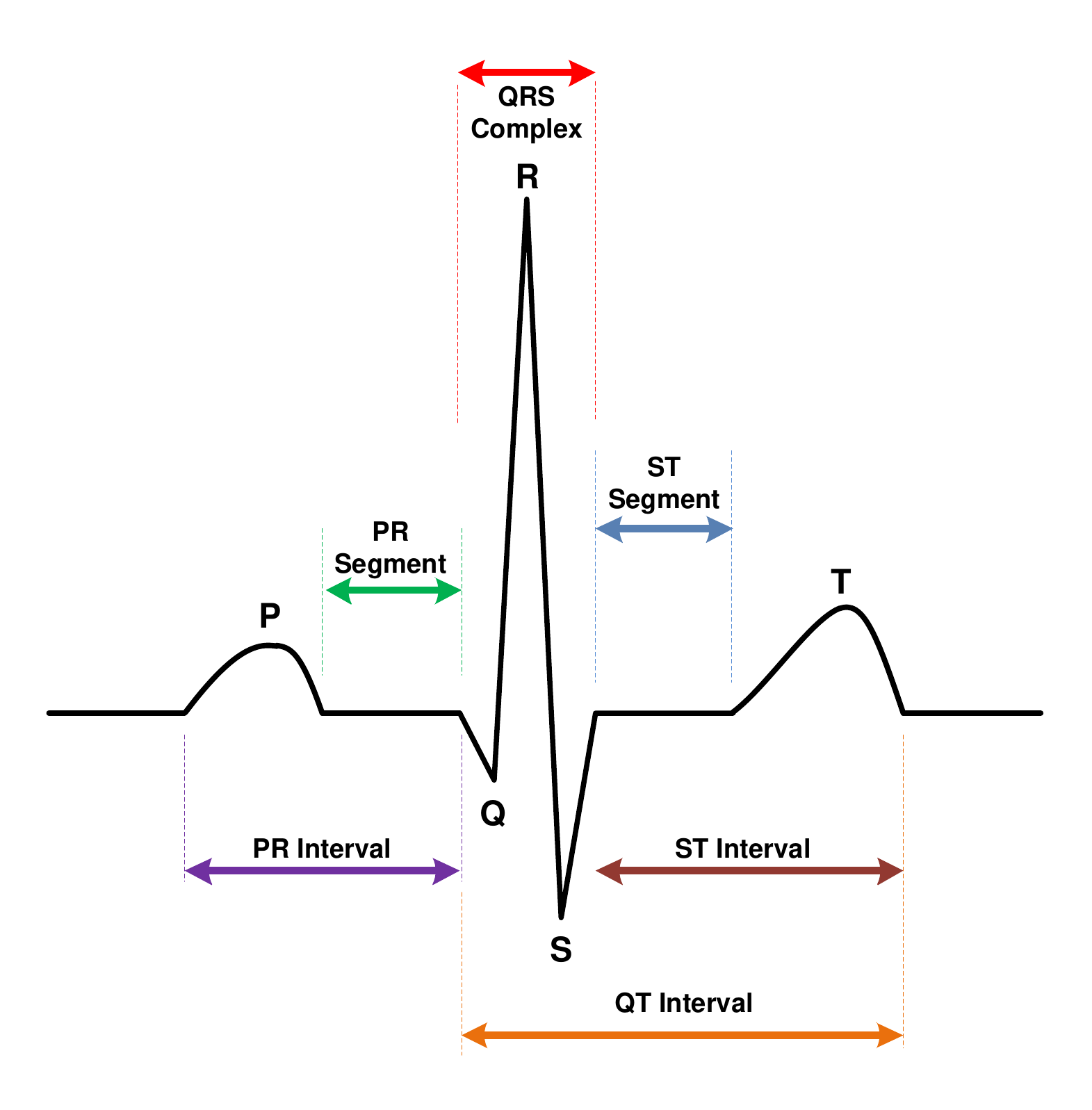}
\caption{A typical heartbeat (one-cycle ECG)} \label{fig:ecg}
\end{figure}

Out of 12 leads, each lead localizes the heart with different viewing angle and gathered wave signals. The recorded ECG signal with 12 leads comprises of five waves represented as $P$, $Q$, $R$, $S$, and $T$. Figure~\ref{fig:ecg} portrays a general one-cycle ECG waveform containing the main interval and segments of the corresponding five waves as discussed. The atrial depolarization generates the $P$ wave, while the $PR$ segment depicts the duration of atrioventricular conduction (AV). $QRS$ is termed as complex labeled and represents the activation of ventricular depolarization. However, $ST-T$ represents the Ventricular repolarization. ECG plays a vital role in the diagnosis and treatment of cardiac abnormalities.Having the early acquisition and correct interpretation of ECG is necessary. However, it is recorded by a professional once the patient arrives at an hospital or using a portable device called Holter ECG that a patient can take at home for 24h or 48h of continuous recording and then bring it back to doctors for insight analysis. Generally, these ECG monitoring approaches are used to record and monitor the ECG data for a short period (usually minutes to hours) and some episodic symptoms probably may not occur during the monitoring period. Furthermore, ECG is generally recorded after encountering the initial symptoms. It is possible that some damage to the heart may already have occurred before reporting to a cardiologist. Therefore, it is deemed necessary to have ECG recording systems which should be capable of recording data remotely with real-time analysis. Keeping in view those as mentioned above it is deemed requirement of real-time monitoring systems for CVDs with the capabilities of early detection of anomalies and alerting the relevant health care providers.\par 
The main contribution of this research work is the detection and classification of CVDs into four classes: Normal, PAC, PVC, and MI. The presented work uses the Discrete Wavelet Transform (DWT) for preprocessing of ECG signal and extracts features using UWT. After filtering and extracting the features, the classification is performed using Bayesian Belief Network. For the removal of false alarms, Tukey box uses temporal consistency analysis.  By extracting features from sliding windows retaining the latest $N$ past values it measures the deviation of the current heartbeat with the vigorous statistical values.\par
The remaining part of the paper is structured as follows: Section~\ref{sec:related} comprehends the related work. Section~\ref{sec:approach} enlightens the proposed approach for CVD detection and classification. Section~\ref{sec:results} depicts the experimental results on real ECG data sets with CVDs. Finally, section~\ref{sec:conclusion} summarizes the research work.\par.  

%%%%%%%%%%%%%%%%%%%%%%%%%%%%%%%%%%%%%%%%%%%%%%%%%%%%%%%
\section{Related work} \label{sec:related}
Numerous vendors and nano-tech industries have introduced different varieties of WBANs and ECG monitoring systems since the last decade. These ECG systems are WBANs are widely in healthcare application for the collection of patient data and transmit to a relevant healthcare service provider.  These devices are advantageous for medical practitioners to collect and monitor their patient's health remotely. After a thorough investigation it is observed that the use of these devices is prevailing in healthcare, but still, many improvements can be incorporated.  These ECG devices are not capable of processing the ECG signals nor having a mechanism to detect and notify the CVD anomaly.  Moreover,  another limitation not capable for the provision of Heart Rate Variability (HRV) analysis.  Those mentioned above are the main limitations existing in the healthcare monitoring systems for instance, 
LifeMonitor~\cite{Equivital}, MyHeart~\cite{luprano2006combination}, CodeBlue~\cite{malan2004codeblue}, LiveNet~\cite{sung2004livenet}, MEDiSN~\cite{ko2010medisn}, AliveCor~\cite{saxon2013ubiquitous}, PhysioMem~\cite{PhysioMem},etc.\par

Lately, a very few approaches have been coined to discourse the gaps and limitations of automatic detection of anomalies and generating alarms for irregular rhythms in respect of WBANs, and real-time ECG recording and monitoring remotely. HeartSaver system is such a kind of system projected in ~\cite{sankari2011heartsaver} with similar intentions.  The system is capable of analyzing the ECG in real-time and can detect cardiac pathologies which are atrioventricular block, atrial fibrillation (A-fib) and myocardial infarction.  The author in ~\cite{ashrafuzzaman2013heart} introduced a mobile application which uses the mobile camera for heart attack detection by placing the index figure on camera.  The proposed approach works only with Heart Rate by calculating the blood peaks.  The purposed approach is unable to detect CVD because CVD detection involves complex features representation and analysis of the ECG signal.\par  

%%%%A new work ~\cite{goodfellow2018towards} discuss this

A wireless-based real-time ECG monitoring system named RECAD was proposed for the detection of an arrhythmia in ~\cite{zhou2006real}. Moreover, another research work has been presented in ~\cite{oresko2010wearable} which uses smartphone platform for the detection of cardiac anomalies in real-time. However, this approach performs CVD classification into paced beat (PACE), PFUS beat, right bundle branch block beat (RBBB) and PVC. In addition to approaches as discussed above, there are other numerous portable and innocuous devices commonly available in markets providing arrhythmia detection amenities, for instance, BodyGuardian~\cite{BodyGuardian}, CardioNet~\cite{cardionet}, and Smartheart~\cite{Smartheart}. However, these devices are unable to detect first hand or initial symptoms relevant to PAC, PVC, and MI.  

Authors in~\cite{jovic2011electrocardiogram} proposed a classification approach based on the analysis of 11 HRV features to distinguish between normal and abnormal ECG. They also further processed and classified into four categories: supra-ventricular arrhythmia, congestive heart failure, arrhythmia, and normal heart-beat. Seven clustering and classification algorithms analyzed ECG records from online databases. As per results reported by the author, they  have three top accurate classification methods for the prediction of binary classes : Random Forest (RF) with 99.7\%, Artificial Neural Network (ANN) with 99.1\% and Support Vector Machines (SVM) with 98.9\% accuracy, and in the case of four classes : SVM with 98.4\%, BNC with 99.4\% ,and RF with 99.6\%.

In~\cite{jadhav2010artificial}, Artificial Neural Network (ANN) classifier is used to detect an arrhythmia in 12-lead ECG data. Authors used UCI arrhythmia dataset for the training of three different classifiers. Their experimental setup attained the accuracy of 86.67\%, and the reported sensitivity was 3.75\%.  Author's first two tested classifiers achieved these results. As per this paper, the third classifier achieved the specificity of 93.1\%. However, the problem with aforesaid approaches is they require high computation power and are not suitable for real-time wireless systems.\par
Another approach was presented for the detection of Coronary Artery Disease (CAD) automatically in ~\cite{giri2013automated}. The proposed approach used four classifiers namely (PNN, KNN, SVM, and GMM) and data dimensionality reduction techniques, for instance, ICA, LDA, and PCA. They have tried several combinations of classifiers and data reduction techniques, and according to their results, ICA with GMM combination was outperforming with the highest accuracy of 96.8\% in comparison of others. 
All the approaches and models reported in this literature review section yielded very good accuracies. However, these models are not feasible for real-time wireless systems due to the requirement of high computational power and resources.\par
In this research work, we have presented a system capable of monitoring and analyzing ECG signal remotely with the blend of supervised machine learning techniques and statistical analysis. 
First, we use signal processing techniques to filter the captured ECG from the noise and extract nine features of heartbeat parameters followed by  Bayesian Network Classifier is used to classify a given captured ECG as normal from the other three anomalies.
\begin{itemize}[leftmargin=*,label={--}]
\item Monitor real-time ECG data and processing it to become feasible for a Wireless ECG monitoring Devices remotely.
\item  Reduction of false alarms using the optimized prediction model.
\item Detection and Prediction of cardiac anomalies: PVC, PAC, and MI.
\end{itemize}
%%%%%%%%%%%%%%%%%%%%%%%%%%%%%%%%%%%%%%%%%%%%%%%%%%%%%%%%%%%%%
\section{Proposed Approach} \label{sec:approach}

Our proposed system is comprised of five layers. The wireless sensors are at first layer which measures ECG and sends the measurements to smartphones in real-time as depicted in Figure ~\ref{fig:architecture}: we have utilized the real development environment for our proposed system for real-time monitoring and transmitting data remotely using these sensors. The brief details of these five layers are given as under: 
\begin{figure}[!htb]
\centering
\includegraphics[scale=0.7]{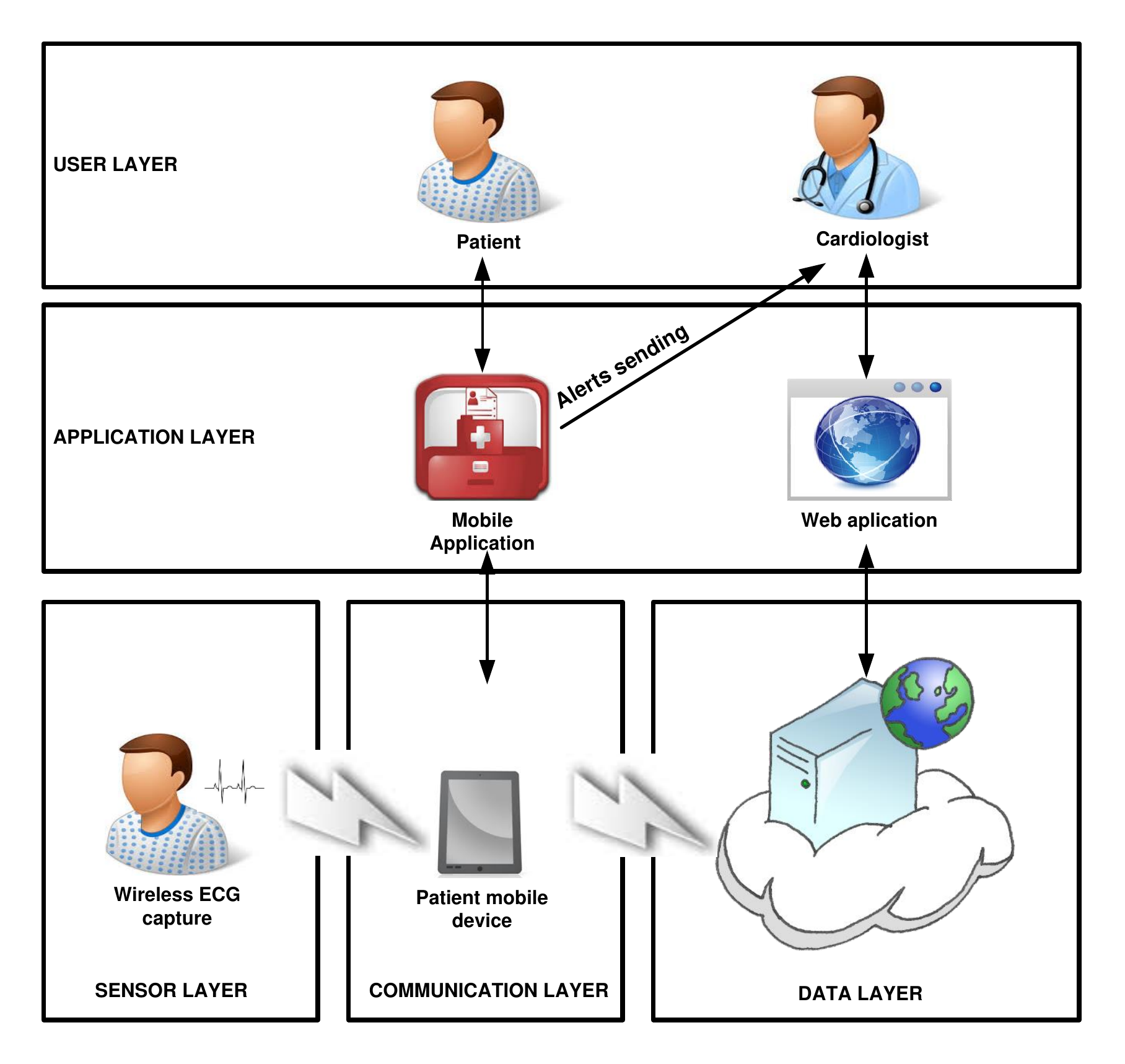}
\caption{Architecture of the proposed system} \label{fig:architecture}
\end{figure}

\begin{itemize}[leftmargin=*,labelsep=5.8mm]
\item Sensor Layer:The sensor layer is the first layer containing the installed ECG sensors with wireless functionality to record, sample, and transmits ECG data. Contrary to the traditional 12-lead ECG system available at hospitals these miniature sensors are manufactured with the intentions of portability, mobility, and specifically designed for smartphone applications. They contain two or three leads for ECG monitoring and sometimes one lead as the latest introduced by Apple. 

\item Communication Layer: This layer uses a smartphone as a gateway to receive and transmit sampled ECG signal by using Bluetooth. The ECG signal captured by these sensors are transmitted along with alarms and notifications to a respective cardiologist or care provider. The transmitted data also contains information about the physical condition and position of the patient and smartphone are prominent because of their activation as a gateway. 

\item Application Layer: represent the mobile application can be developed in IOS and Android for smartphones. The application and smartphone resources are utilized for the preprocessing and extraction of ECG signal. Based on extracted features this application detects the cardiac anomalies for the further expert opinion.
\item Data Layer: this layer contains the EHR data and deals with the connectivity of databases.  The associated healthcare databases reside here which are responsible for the storage and retrieval of patient data. The data constitutes of patient medical history, present illness state, general information, ECG captured, alert, and extracted parameters. 
\item User Layer: this layer provides the end user interface on a smartphone application. The layer is mainly centric on patients and healthcare providers. This layer is responsible for receiving the alerts and CVD detections which provoke the cardiologist to take appropriate action of medical assistance to his patient according to need.
\end{itemize}

\begin{figure}[!htb]
\centering
\includegraphics[scale=0.70]{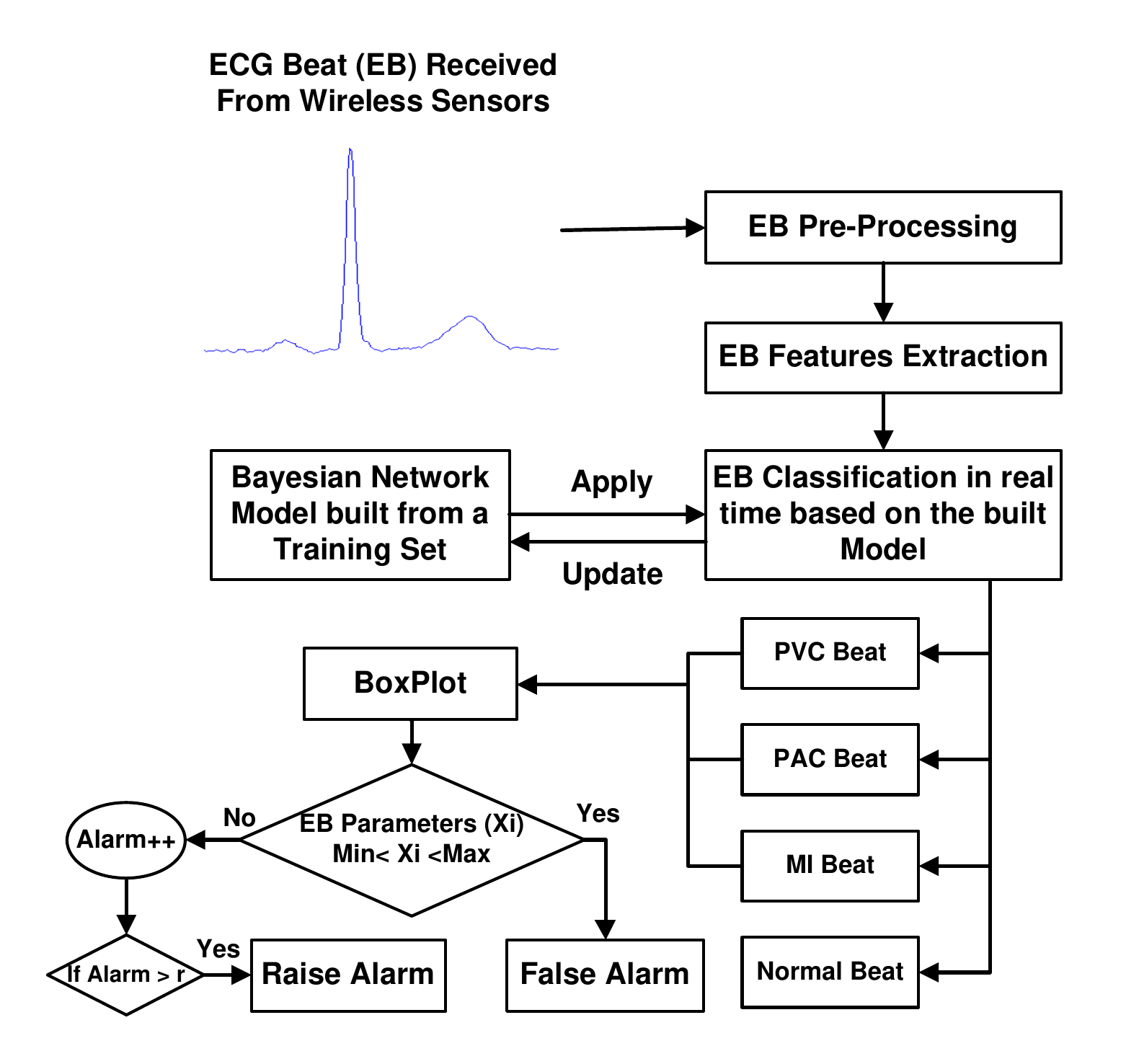}
%\captionsetup{justification=centering}
\caption{Design steps of the application layer} \label{fig:design}
\end{figure}

Our approach mainly revolves around the application layer. 
The main contribution of our proposed work comprises of various steps involving the preprocessing of sampled ECG using DWT, extraction of features from the ECG signal ($P$, $Q$, $R$, $S$, $T$) using UWT and perform analysis and prediction for CVD on these extracted features with the blend of Machine learning using supervised classification method a temporal analysis. A probabilistic model named Bayesian Network is used for the detection of cardiac anomalies and the prediction of MI, PVC, and PAC from a regular heartbeat.  A dual verification and confirmation system is designed to check the validity of detected cardiac anomaly using Tukey box analysis.  The purpose of using the Tukey box system as a wrapper on Bayesian Network is to make it more robust,  minimize the false alarms and produce authenticity for prediction and detection. Bayesian Network is trained on real ECG data and functionality is added to make it reliable and adaptive to combat new ECG features. The flowchart of the proposed approach is given in Figure ~\ref{fig:design} \par

%%%%%%%%%%%%%

\subsection{Pre-Processing}\label{sec:processing}
Pre-processing is required to eradicate the signal to noise ratio from the 
ECG signal acquired from WBAN sensors. The existing noises interfere with the cardiac equipment installed in the frequency range of [0.01-150]Hz. The most common noise signals which cause interference are:
\begin{itemize}[leftmargin=*,label={--}]
\item Electromyographic (EMG) noise signals range from 25Hz to 100Hz.
\item Electrodes gesticulation artifacts vary between 1Hz to 10Hz
\item Baseline wandering with a frequency below 0.5Hz, this wandering is caused by the patient’s body parts movement and natural respiration process. This noise dispositions the iso-electric line of the measured ECG). It can be suppressed without loss in the original signal, by a high-pass digital filter or by the use of a standard Wavelet Transform (WT). In this study, we use a 0.5Hz FIR high-pass digital filter (Finite Impulse Response).
\item Power line noise: Electronic circuits of ECG sensors generate these kinds of noises. 
\end{itemize} \par
Among the above mentioned, Baseline wondering and Power line interference are two primary noise signals which degrade detection accuracy of the ECG features substantially, in particular, it impacts QRS complex badly. Other noises may be wide-band and usually complex stochastic processes which distort the ECG signal. The other noises including EMG and electrode motion artifacts are most difficult signals to eradicate. Traditional digital filtering schemes are unable to remove these noises because they involved some complex probabilistic schemes which can cause potential interference with ECG frequency. Oppose to traditional digital filters we have used one of the well-known method named Wavelet Transform (WT) for time-frequency transformation. For preprocessing of data, we have applied Discrete Wavelet Transform (DWT) to tackle the discrete nature of ECG signal. DWT is applicable in various domains due to its property of associating the frequency signals and timestamps. , especially for denoising. There are two main types of WT known as Discrete Wavelet Transform (DWT) and Continuous Wavelet Transform (CWT). For preprocessing of data, we have applied Discrete Wavelet Transform (DWT) to tackle the discrete nature of ECG signal.\par
\begin{figure}[!htb]
\centering
\includegraphics[scale=0.6]{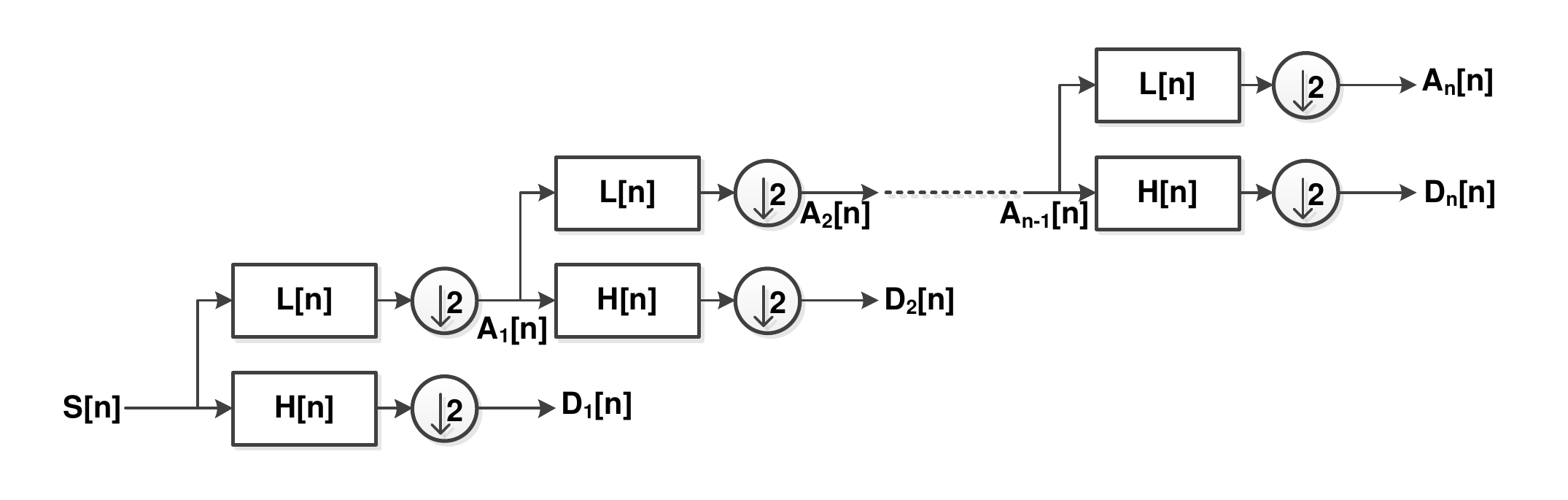}
\caption{Discrete Wavelet Decomposition} \label{fig:dwt}
\end{figure}

 
In DWT, the signal $S$ is convolved and dislodge by passing it through a series of filters. Concurrently this signal $S$ is passed via a high pass filter $H$ and low pass filter $L$. The output of these filters yielded approximated coefficient $A$ for low pass filter and detailed coefficient $D$ in respect of high pass filter. This output is further sub-sampled by the factor two which eradicates half of the frequencies from each output as specified in the following equation. 
\begin{equation}\label{eq1}
{s_L}[n] = {A_1}[n] = \sum\limits_{k =  - \infty }^\infty  {s[k]L[2n - k]}
\end{equation}
%%%%%%%%%%%%%%%%%%%%%%%%%%%%%%%%
\begin{equation}\label{eq2}
{s_H}[n] = {D_1}[n] = \sum\limits_{k =  - \infty }^\infty  {s[k]H[2n - k]}
\end{equation}
%%%%%%%%%%%%%%%%%%%%%%%%%%%%%%%%%%%%
For the further decomposition of the yielded approximation coefficient, it is further decimated through a series of low pass filters and high pass filters as shown in the Figure~\ref{fig:dwt}. Lastly, reconstruction of a denoised signal is performed using wavelet function. The linear combination of wavelet functions weighted by wavelet coefficients is used to achieve signal reconstruction. \par
%%%%%%%
%%%%%%%%
The classical DWT does not exhibits shift invariant property; this means that DWT of a translated version of a signal $S$ is not the same as the DWT of the original signal, and shift-invariance property is vital for denoising and pattern recognition applications.
In order to overcome this drawback and get complete characteristic of the original signal, we use a variant of DWT named Undecimated Wavelet Transform (UWT). The quality of this method is that it does not decimate the signal, while it offers a better balance between smoothness and accuracy than the DWT. 
We have applied the same DWT technique we have implemented in our previous work related to cardiac anomalies  ~\cite{hadjem2014ecg}. In DWT, the decimation retains even indexed elements. However, the decimation could be carried out by choosing odd indexed elements instead of even indexed elements. This choice concerns every step of the decomposition process, so at every level, we chose odd or even. If we perform all the different possible decompositions of the original signal, we have 2J different decompositions, for a given maximum level J.
Let us denote by $\epsilon$j = 1 or 0 the choice of odd or even indexed elements at step j. Every decomposition is labeled by a sequence of 0s and 1s: $\epsilon = \epsilon1...,\epsilon$j. This transform is called the $\epsilon$-decimated DWT as given in our previous work~\cite{hadjem2014ecg}.  The basis vectors of the $\epsilon$-decimated DWT can be obtained from the standard DWT. The further details about DWT and its applicability in various domains are comprehensively explained in ~\cite{wavelet}.\par
UWT first decomposes the ECG signal into several sub-bands by applying the Discrete Wavelet Transform and then modifies each wavelet coefficient by applying a threshold function, and finally reconstructs the de-noised signal. This technique ensures no loss of signal sharpest features by discarding only the portions of the details that exceed a certain limit due to the occurrence of noise. 
In this research work, we have employed UWT for both feature extraction and pre-processing of signals. The further detailed explanation of UWT applicability in this research is explained in next sub-section.\par
%%%%%%%%%%%%%%% End of Sub section%%%%%%%%%%%%%%%%
\subsection{Features Extraction}\label{sec:features}

Our proposed technique for feature extraction is inspired by Boosting (to conquer the complex data points first)  which is a supervised learning technique. After preprocessing and denoising, like boosting this technique detects all the complex peaks, i.e.,$QRS$ . After extracting the features from complex peaks $QRS$, it opts for other relatively simple waves including $P$ and $T$. It is imperative to mention that feature extraction of all other waves is extracted from the original ECG signal except the $QRS$ complex. For the extraction of complex $QRS$ peaks, we are using the pre-processed ECG signal because accurate detection of complex $QRS$ is cumbersome from the original signal. Our feature extraction algorithm is inspired by ~\cite{bhyri2009estimation}. We have used the  DWT variant named UWT for feature extraction. The rationale behind UWT usage the maintainability of the complete characteristics of the original signal as discussed in ~\cite{nason1995stationary}.\par

\begin{figure}[!htb]
\centering
\includegraphics[scale=0.7]{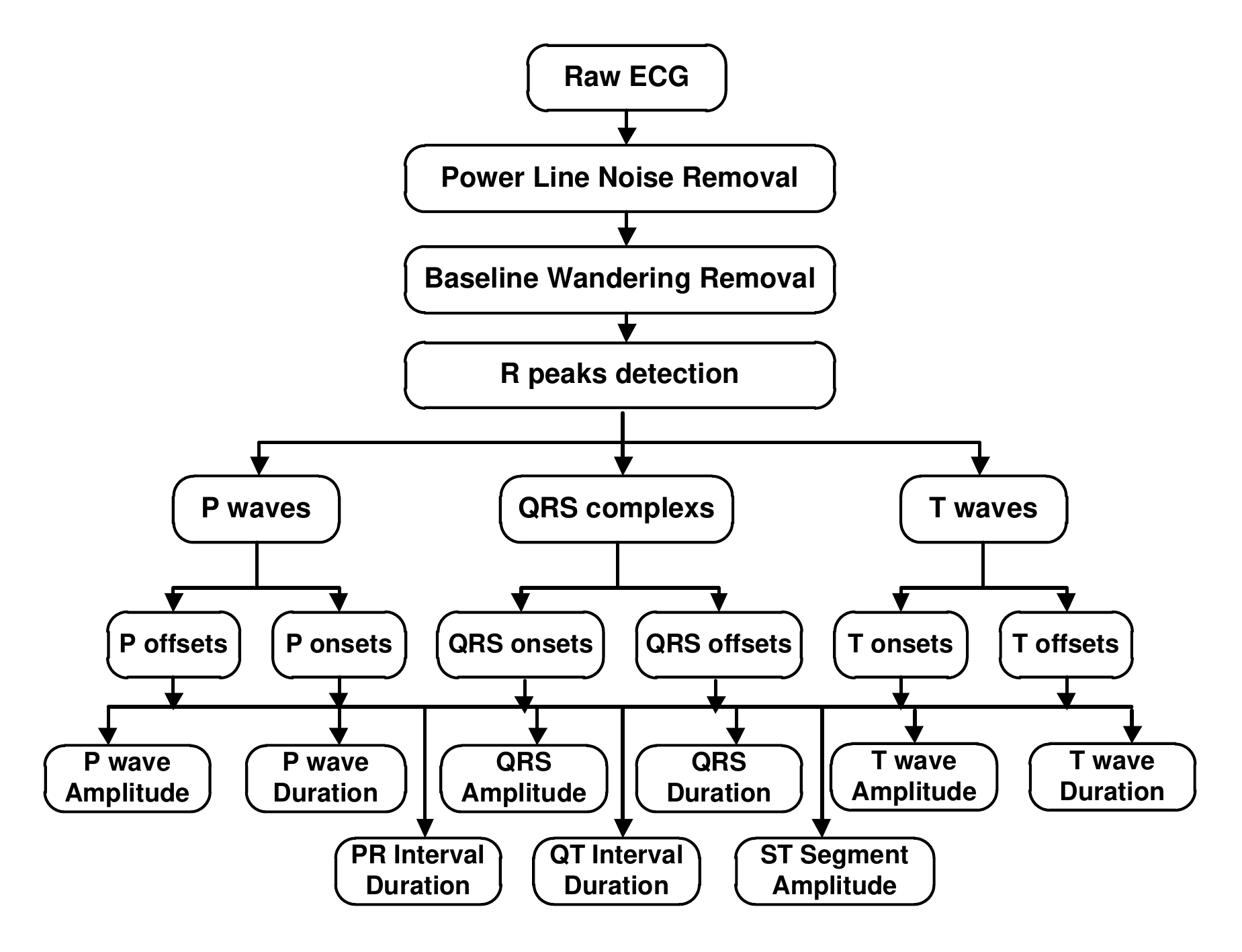}
\caption{ECG signal processing and features extraction} \label{fig:features}
\end{figure}

UWT is applied using the Daubechies wavelet to decompose the ECG signal into eight levels in the first step. In the second step, the reconstruction of the signal is completed using detail and approximation coefficients of all frequency bands. The next step is the detection of peaks in which the algorithm detect $P$, $QRS$ complex, and $T$ peaks by adequately specifying a threshold. This threshold is used for the acceptance and rejection of peaks of a particular amplitude. Moreover, the algorithm uses a window width for the detection of onsets and offsets by specifying the current number of samples of the signal. The high-level diagram of feature extraction is given in Figure \ref {fig:features}.
%%%%%%End Figure%%%

%%%%Figure%%%
\par 
The algorithmic description of the peak detection for $QRS$, $P$, and $T$ are enumerated as given:

\begin{itemize}[leftmargin=*,labelsep=5.8mm]

\item Apply Daubechies wavelet for eight levels UWT on ECG  input signal.
 \item Detection of zero crossing points in detailed coefficients at all levels yielded by high pass filter.
\item For the coarse estimation of real peaks, the zero crossing point is set to the large scale. 
\item  For each detected point, algorithm searches for the fine scale for the nearest zero point
\item Repeat the last step until it obtains the optimal scale.
\end{itemize}
In order to obtain the zero crossing points for calculating the offsets and onsets, a time window is used. The optimal window size we have used is 50 milliseconds.  Based on the detection of  $QRS$ multiple peaks, the signal is scanned for 50 milliseconds on the left side to get the minimum value, and this corresponds to Q peaks. Similarly, the minimum value on the right side of the R peaks corresponds to "S" peaks. In a specific case if a particular signal is not crossing the zero line, then we consider the minimum value inside that particular window as offsets and onsets. This mechanism results in the extraction of six features from each one-cycle ECG beat as the first six feature mentioned in Table ~\ref{table:ecg_parameters}.\par  It is relatively easy to calculate the amplitude, segment, duration of each ECG wave, and intervals from the nine features as presented in Table~\ref{table:ecg_parameters}. Moreover, it is mandatory to highlight that these nine features are generally of high interest and high importance for a cardiologists ~\cite{morris2009abc} 
%%%%%%%%%%%%%%%%%%%%%%%%%%%%%%%%%%%%%%%%%%%%%%%%%%%%%%%
\begin{table}[H]
\begin{tabular*}{\textwidth}{|p{1.5cm}|p{13.2cm}|@{\extracolsep{\fill}}}
\toprule
  \hline
   \textbf{Parameter} & \textbf{Description} \\
  \hline
  \midrule
	\textbf{$P_{amp}$} & Amplitude in \textit{mV} of the P wave calculated by searching the peak between Ponset and Poffset \\
	\hline
	\textbf{$P_{dur}$} & Duration in \textit{seconds} of the P wave between Ponset and Poffset \\
	\hline
	\textbf{$QRS_{amp}$} & Amplitude in \textit{mV} of the QRS wave calculated by searching the peak between QRSonset and QRSoffset \\
	\hline
	\textbf{$QRS_{dur}$} & Duration in \textit{seconds} of the QRS wave between QRSonset and QRSoffset \\
	\hline
   \textbf{$T_{amp}$} & Amplitude in \textit{mV} of the T wave calculated by searching the peak between Tonset and Toffset \\
	\hline
   \textbf{$T_{dur}$} & Duration in \textit{seconds} of the T wave between Tonset and Toffset \\
  \hline
	\textbf{$PR_{dur}$} & Duration in \textit{mV} of the PR Interval between Ponset and QRSonset \\
	\hline
	\textbf{$ST_{amp}$} & Amplitude in \textit{mV} of the ST segment calculated based on local maxima between QRSoffset and Tonset \\
	\hline
	\textbf{$QT_{dur}$} & Duration in \textit{seconds} of the QT interval between QRSonset and the Toffset \\
	\hline
	\bottomrule
\end{tabular*}
\caption{9 parameters calculated for each ECG Beat}
\label{table:ecg_parameters}
\end{table}

%\begin{table}[H]
%\centering
%\begin{tabular}{|c|c|}
%\toprule
 % \hline
  % \textbf{Parameter} & \textbf{Description}   \\
  %\hline
  %\midrule
%	\textbf{$P_{amp}$} & Amplitude in \textit{mV} of the P wave %calculated by searching the peak between Ponset and Poffset \\
%	\hline
%	\textbf{$P_{dur}$} & Duration in \textit{seconds} of the P wave between Ponset and Poffset \\
	%\hline
%	\textbf{$QRS_{amp}$} & Amplitude in \textit{mV} of the QRS wave calculated by searching the peak between QRSonset and QRSoffset \\
%	\hline
	%\textbf{$QRS_{dur}$} & Duration in \textit{seconds} of the QRS wave between QRSonset and QRSoffset \\
%%	\hline
  % \textbf{$T_{amp}$} & Amplitude in \textit{mV} of the T wave calculated by searching the peak between Tonset and Toffset \\
%	\hline
  % \textbf{$T_{dur}$} & Duration in \textit{seconds} of the T wave between Tonset and Toffset \\
  %\hline
%	\textbf{$PR_{dur}$} & Duration in \textit{mV} of the PR Interval between Ponset and QRSonset \\
	%\hline
	%\textbf{$ST_{amp}$} & Amplitude in \textit{mV} of the ST segment calculated based on local maxima between QRSoffset and Tonset \\
	%\hline
	%%\textbf{$QT_{dur}$} & Duration in \textit{seconds} of the QT interval between QRSonset and the Toffset \\
	%\hline
	%\bottomrule
%\end{tabular} 
%\caption{9 parameters calculated for each ECG Beat}
%\label{table:ecg_parameters}
%\end{table}

%%%%%%%%%%%%%%%%%%%%%%%%%%%%%%%%%%%%%%%%%%%%%%%%%%%%%%%%%
These nine extracted features are used as input variables to BNC for CVD detection. The details are given in the following subsection.

%%%%%%%%%%%%%%%%%%%%%%%%%%%%%%%%%%%%%%%%%%
\subsection{Bayesian Network Classifier (BNC)}\label{sec:bnc}

A Bayesian Network is a machine learning technique based on statistical probabilities.  The famous Bayesian theorem is the backbone of a Bayesian Network which represents a set of random variables and their conditional dependencies.  A Directed Acyclic Graph (DAG) usually represents the conditional properties and random variables that's why it is also known as a probabilistic directed acyclic graphical model. In bioinformatics, these Bayesian networks are instrumental in showing the symptoms and diseases cardinalities.  Based on the given symptoms and their conditional probabilities the network can determine the presence of a particular disease. The nodes in the Bayesian Network represents the variables which may be observable quantities or unknown parameters. The edges in the network show the conditional dependencies. Moreover, non-connected nodes represent the observations or variables which are conditionally independent to each other. In the network, there is a probability function associated with each node. The function takes the input of a particular set of values from the parent node and returns the probability of a variable expressed by the node.  There is another variant of BNC called Dynamic BNC which is used to model the succession of variables.  In our methodology, the extracted nine features by DWT and UWT are the node of BNC.\par
%as shown in Figure~\ref{fig:DAG}.\par
%%%%%%%%%%%%%%%%%%%%%%%%%%%%%%%%%%%%%
%\begin{figure}[!htb]
%\centering
%\hspace*{-1cm}\includegraphics[scale=0.5]{./Definitions/figures/pdf/DAG.pdf}
%\caption{Generated DAG by the Bayesian Network} \label{fig:DAG}
%\end{figure}
%%%%%%%%%%%%%%%%%%%%%%%%%%%%%%%%%%%%%%%%%%%%%%%%%%%%%%%%%%%%%%%%%%%%
Let $G = {\rm{ }}\left( {H,E} \right){\rm{ }}$ is a DAG, where E represents the edges and $H$ are the node and let $X = {\rm{ }}{\left( {{X_h}} \right)_{h \in H}}$ a set of variables indexed by $H$ as mentioned in our previous work ~\cite{hadjem2014ecg}. Moreover,$X$ represents a Bayesian Network, and its combined probability density function can be expressed by a product of the individual density functions and conditional dependencies on their parent variables as given in Equation \ref{eq10}.\par
%%%%% EQ%%%
\begin{equation}\label{eq10}
p(x) = \prod\limits_{h \in H} p \left( {{x_h}{\mkern 1mu} |{\mkern 1mu} {x_{{\mathop{\rm pa}\nolimits} (h)}}} \right)
\end{equation}
%%% EQ%%%
where in Equation \ref{eq10}  pa ($h$) refered to a parent set for $h$.

In order to calculate the conditional probability for any member of a joint distribution for a given set of random variables, following chain rain can be used:
%%%% EQ%%%%%
\begin{multline}\label{eq3}
{\rm{P}}({X_1} = {x_1}, \ldots ,{X_n} = {x_n}) =
\prod\limits_{h = 1}^n {\rm{P}} \left( {{X_h} = {x_h}\mid {X_{h + 1}} = {x_{h + 1}}, \ldots ,{X_n} = {x_n}} \right)
\end{multline}
%%%%%%%

One can refer ~\cite{Bayesian_Approch} to get thorough understanding and knowledge about the applications of Bayesian network in various domains.

\subsection{Box-and-Whisker plot}\label{sec:box}

\begin{figure}[!htb]
\centering
\includegraphics[scale=0.55]{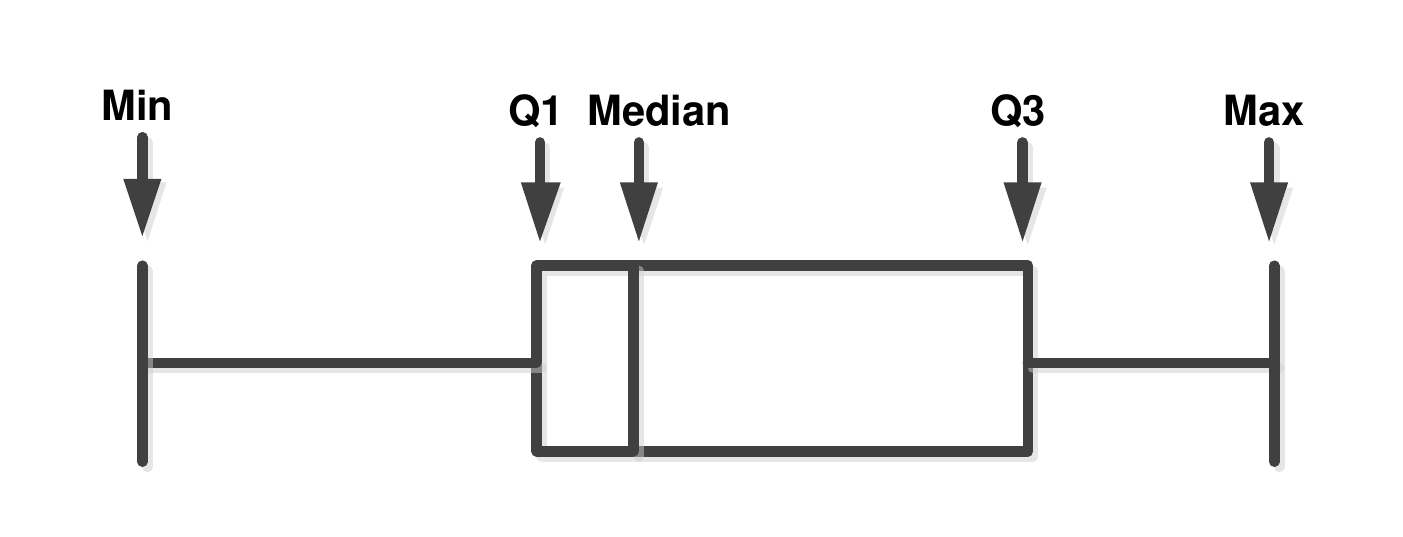}
\caption{Boxplot} \label{fig:boxplot}
\end{figure}

The Box and Whisker plot is a statistical method commonly used to detect and represents the outlier in a given dataset or observations and also known as a boxplot. In order to detect the abnormal measurement, let $X_{i}^{w}=\{x_{i,t-w},\ldots,x_{i,t}\}$ depicts a temporal sliding window of the last $w$ values in respect of $i^{th}$ recorded or monitored ECG parameter. The lower quartile ($Q_1$ is the 25\% and the upper quartile ($Q_3$ is the 75\%) of $X_{i}^{w}$ are capitalized to get robust measures for the mean $\hat\mu =(Q_{1}+Q_{3})/2$, and the standard deviation is replaced by the interquartile range $\hat\sigma= IQR= Q_{3}- Q_{1}$ . The visual representation is specified in the Figure ~\ref{fig:boxplot}.\par    
%$\alpha$ is a parameter which in our case is equal to $1.5$.\par
%%%%%%%%%%%%%%%%%%%
%%%%%%%%%%%%%%%%%%%%%%% FIG %%%%%%%%%%%%
In our proposed approach we have used the training set containing Normal ECG measures/beats to determine the mean and variance as a reference in respect of nine extracted parameters of each ECG Beat ($P_{amp}$, $P_{dur}$, $QRS_{amp}$, $QRS_{dur}$, $T_{amp}$, $T_{dur}$, $PR_{dur}$, $ST_{amp}$, $QT_{dur}$). As mentioned in the earlier subsection, our approach uses Bayesian Network model to detect and classify the abnormality (MI, PVC or PAC) for each ECG Beat. It is imperative to mention that according to medical literature, some parameters are more likely to be abnormal for certain ECG anomalies than others, for example, the ST segment elevation for MI.
After the classification/detection of Bayesian Network for abnormal (PVC, PAC or MI) we apply the univariate boxplot on the extracted nine features of each ECG Beat. If the boxplot confirms the abnormality only then the alarm variable, i.e., $AlarmClass$, corresponding to each anomaly class (PVC, PAC, MI) get incremented and considered. Each ECG Beat contains nine parameters, and if the deviation is detected in at least one parameter and  Boxplot validates it then is treated as abnormal. If no deviation is detected in all nine parameters, then it is considered as Normal ECG Beat regardless of Bayesian prediction and detection. In this scenario, the output of the Bayesian Network is treated as a false alarm and will not be considered. Our system generates the alarm when the value of $AlarmClass$ variable is greater than $r$, we have used a value of $r \ge 3$ in our implementation.
Algorithm ~\ref{alg:alg1} specifies the complete steps of our proposed approach.\par
%%%%%%%%%%%%%%%%%%%%%%%%%%%%%%%%%%%%
\begin{algorithm}[!htb]
\caption{CVD Prediction Algorithm including Preprocessing of ECG signal and its classification}
\label{alg:alg1}
\begin{algorithmic}[1]
\State Build a BN classification model using training set $TS$\\
\State Calculate boxplot parameters for Normal ECG Beats in the $TS$\\
\State Set the windows size $win$, $i=0$\\			
\While {Captured ECG Beat $EB$}	\\										
\State Remove Noise $EB$ with 0.5Hz High-Pass filter and DWT; \\       	
\State Extract ${P,QRS,T}$ Peaks, Onsets and Offsets using UWT;\\
\State Calculate the nine $EB$ features;\\
\If{$i = win$}\\
\For {each $EB$ in the windows $win$}\\
\State Apply the trained BN Model on the nine extracted parameters $X_i$ of the $EB$;\\
\If { $Min < X_i < Max$ for all $i \in 1,2, ...9$ } \\	
\State Normal $EB$, False alarm;\\
\Else\\	
\State Abnormal $EB$;\\
\State $AlarmClass++$;\\
\EndIf\\							
						\EndFor\\
						\If {$AlarmClass > r$}\\
							\State Generate an alarm;\\
						\EndIf\\
						\State Update $TS$ (training Set) and the model with the EB data of the last windows $win$;\\
						\State $i=0$;\\
						\State $AlarmClass=0$;\\
					\Else\\
							\State $i=i+1$;\\							
				  \EndIf\\
				\EndWhile			

\end{algorithmic}
      \end{algorithm}
%%%%%%%%%%%%%%%%%%%%%%%%%%%%%%%%%%%%%%%%%%%%%%%%%%%%

%%%%%%%%%%%%%%%%%%%%%%%%%%%%%%%%%%%%%%%%%%%%%%%%%%%%%%%%%%%%%%%%%%%%%%%%%%%%%%%%
\section{Experimental results} \label{sec:results}

The detail of datasets we have used for the evaluation of our proposed approach is discussed as under:
Two datasets are used from the  Physionet~\cite{PhysioNet}. The first one is the European ST-T Database (EDB)~\cite{EDB}. This dataset contains ECG recordings, and each record comprises of two-hour duration. Each record is recorded with two leads (different for each record) and sampled at   250 samples per second. Moreover, this dataset consists of 90 annotated excerpts of ambulatory ECG recording from 79 subjects. We have used this European ST-T Database (EDB) to get the records containing ECG Beats with Myocardial Infarction (MI) and Normal ECG Beat.  

\begin{figure}[!htbp]  %width=2.5in
\centering
\parbox{0.45\textwidth}{
\subfigure[Normal ECG]{\includegraphics[scale=0.43]{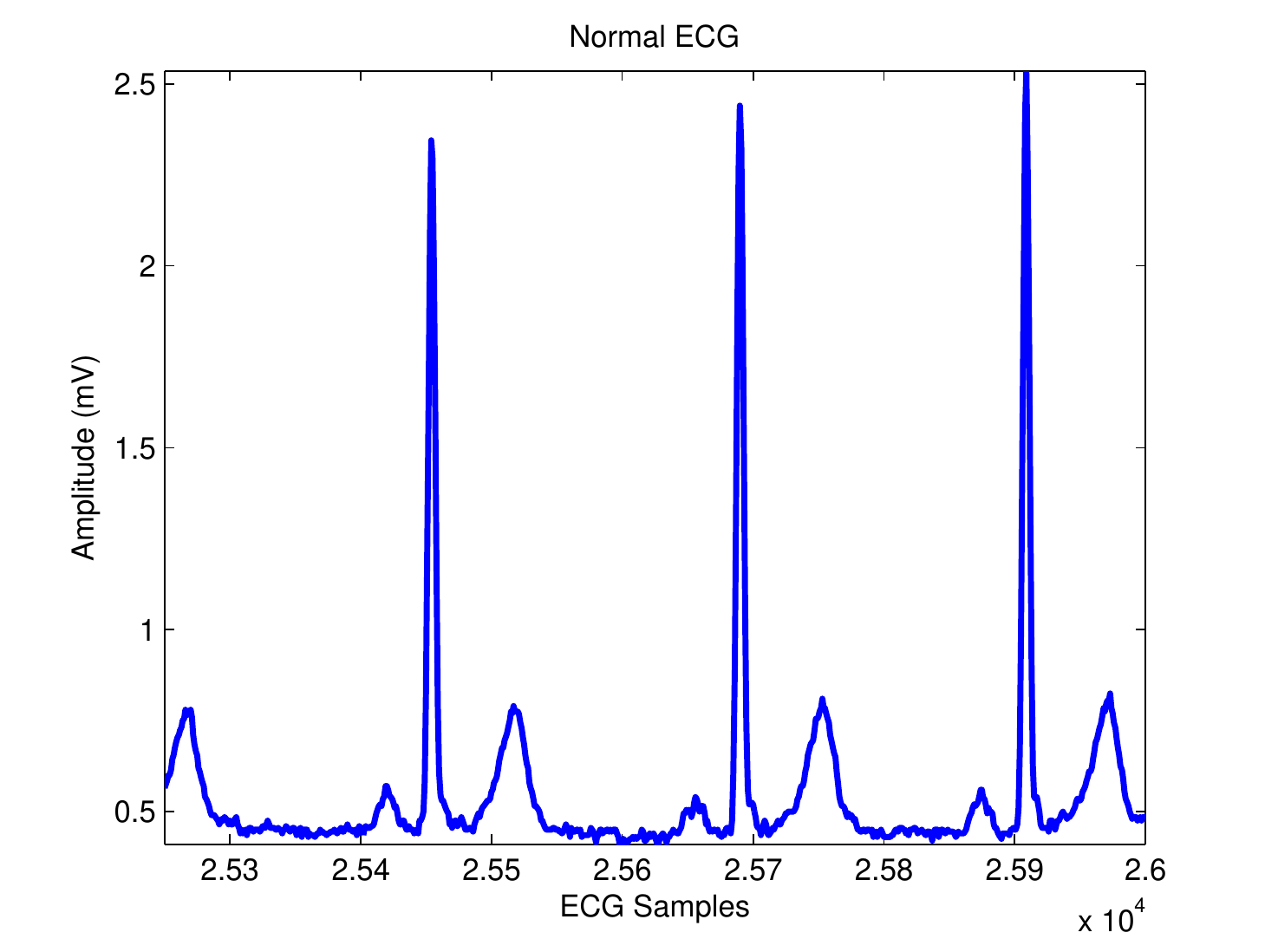}
\label{fig:Normal_ecg}}}
\parbox{0.45\textwidth}{
\subfigure[MI ECG]{\includegraphics[scale=0.43]{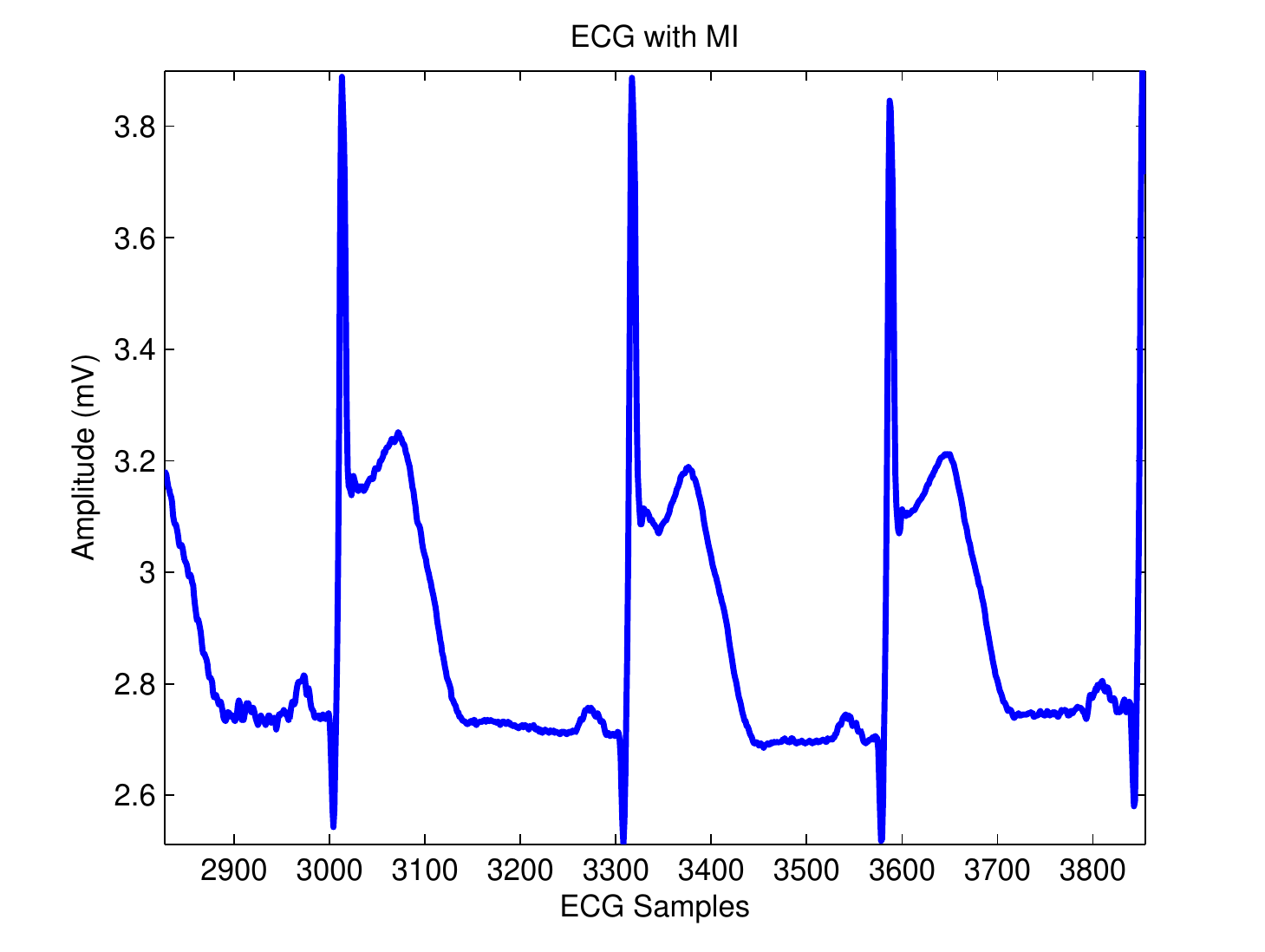}
\label{fig:MI_ecg}}
}
\parbox{0.45\textwidth}{
\subfigure[PVC ECG]{\includegraphics[scale=0.43]{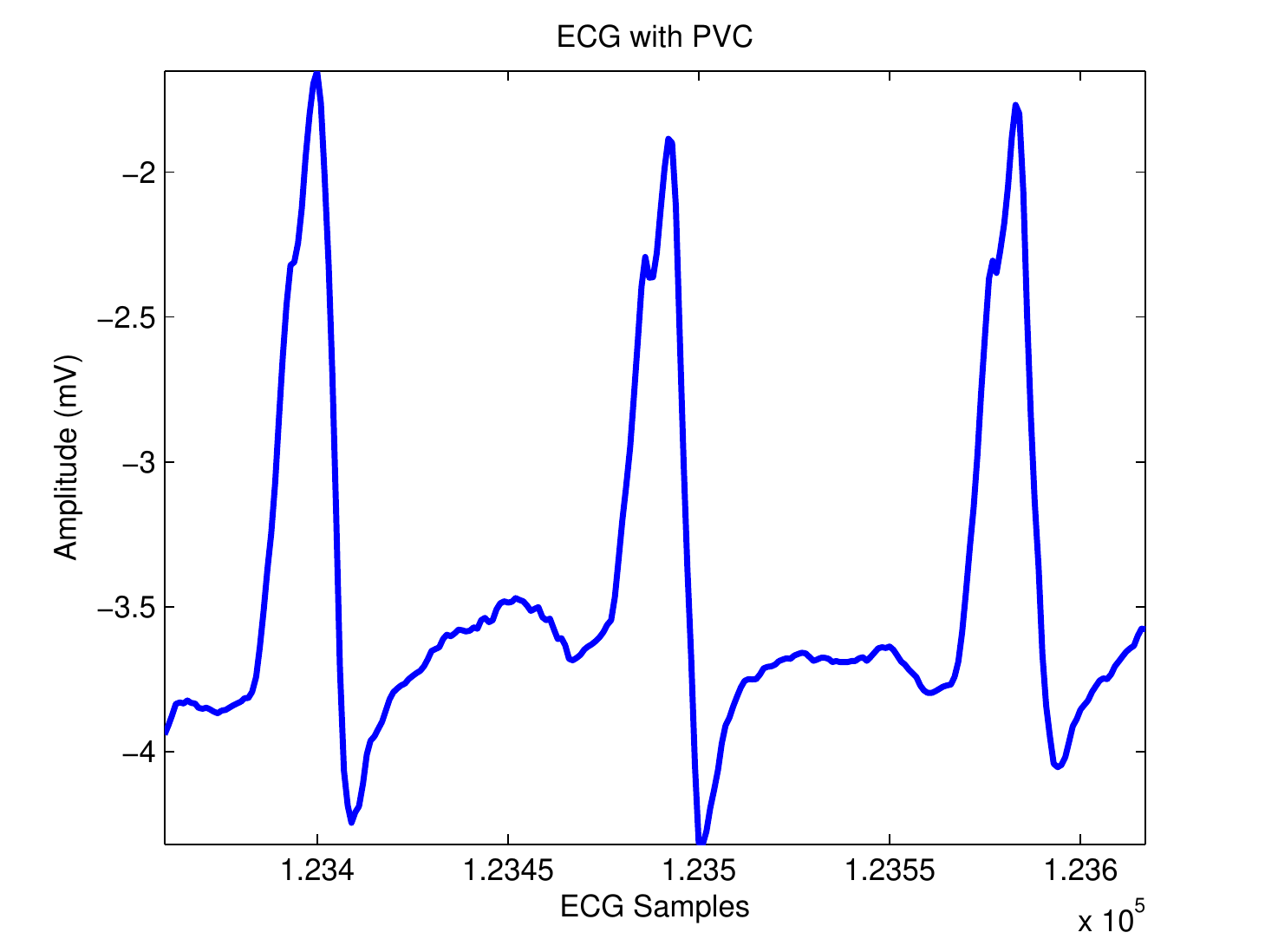}
\label{fig:PVC_ecg}}
}
\parbox{0.45\textwidth}{
\subfigure[PAC ECG]{\includegraphics[scale=0.43]{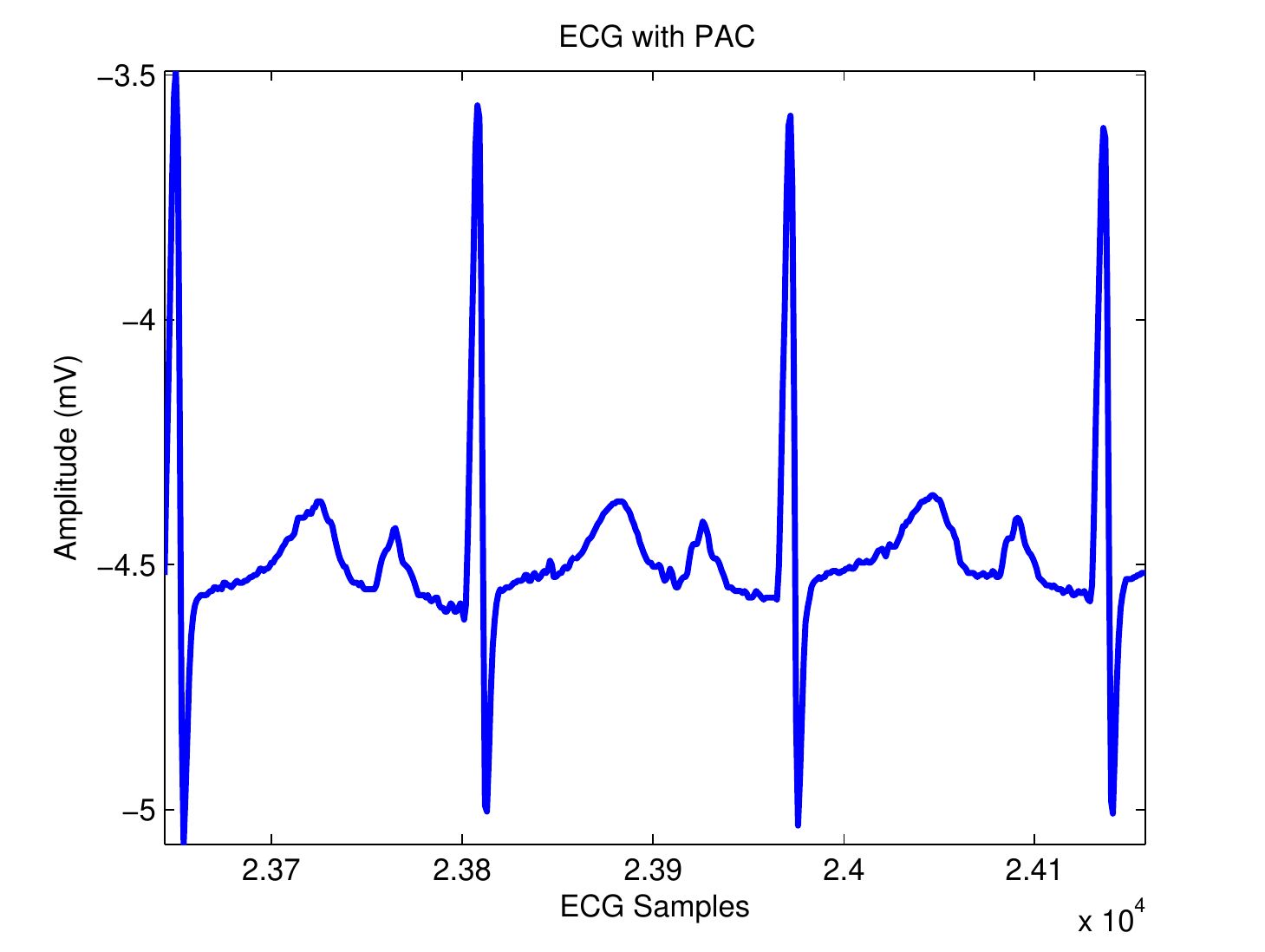}
\label{fig:PAC_ecg}}}
\caption{Various ECGs from Physionet DB}
\label{fig:various_ecg}
\end{figure}

The second dataset is the St. Petersburg Institute of Cardiological Technics 12-lead Arrhythmia Database (INCARTDB)~\cite{PhysioNet}. ECG beats are recorded with traditional 12 leads, and each record is sampled at 257 Hz. This dataset consists of 75 annotated excerpts, and each record comprises of 30 mint duration. These 75 annotated excerpts are extracted from 32 Holter records. We have used this dataset to obtain ECGs with PAC, PVC and Normal ECG. The third dataset is  MIT-BIH Arrhythmia Database (MITDB)~\cite{MITDB} used for various ECGs. This database contains 48 half-hour excerpts of two-channel ambulatory ECG recordings, obtained from 47 subjects; the recordings are digitized at 360 samples per second.
Figure~\ref{fig:various_ecg} is depicting the various ECG signals from Physionet dataset including a normal and MI ECG from EDB. MI ECG is characterized by an elevation of the ST Segment as shown in the figure. We can also see a PVC and PAC ECG from INCARTDB.\par
%%%%%%%%%%%%%%%%%% Figures for Understaning from Physionet%%%%%%%
%%%%%%%%%%%%%%%%%%%%%%%%%%%%%%End Figure%%%%%%%%%%%%%%%%%
%%%%Pre-processed details%%%%%%%%%%%

\begin{figure}[!htb]  %width=2.5in
\centering
\parbox{0.45\textwidth}{
\subfigure[Normal ECG]{\includegraphics[scale=0.43]{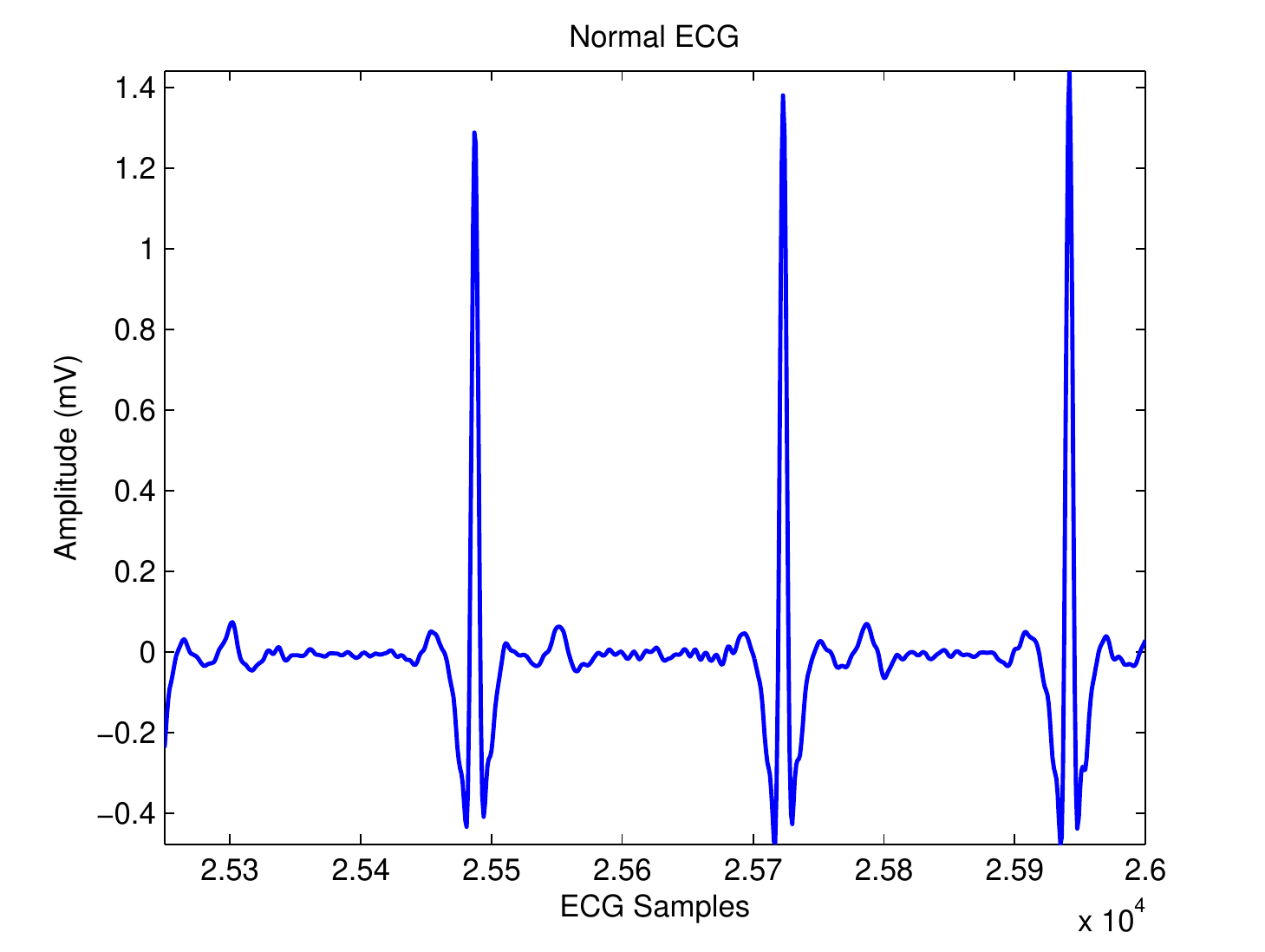}
\label{fig:Normal_ecg_processed}}}
\parbox{0.45\textwidth}{
\subfigure[MI ECG]{\includegraphics[scale=0.43]{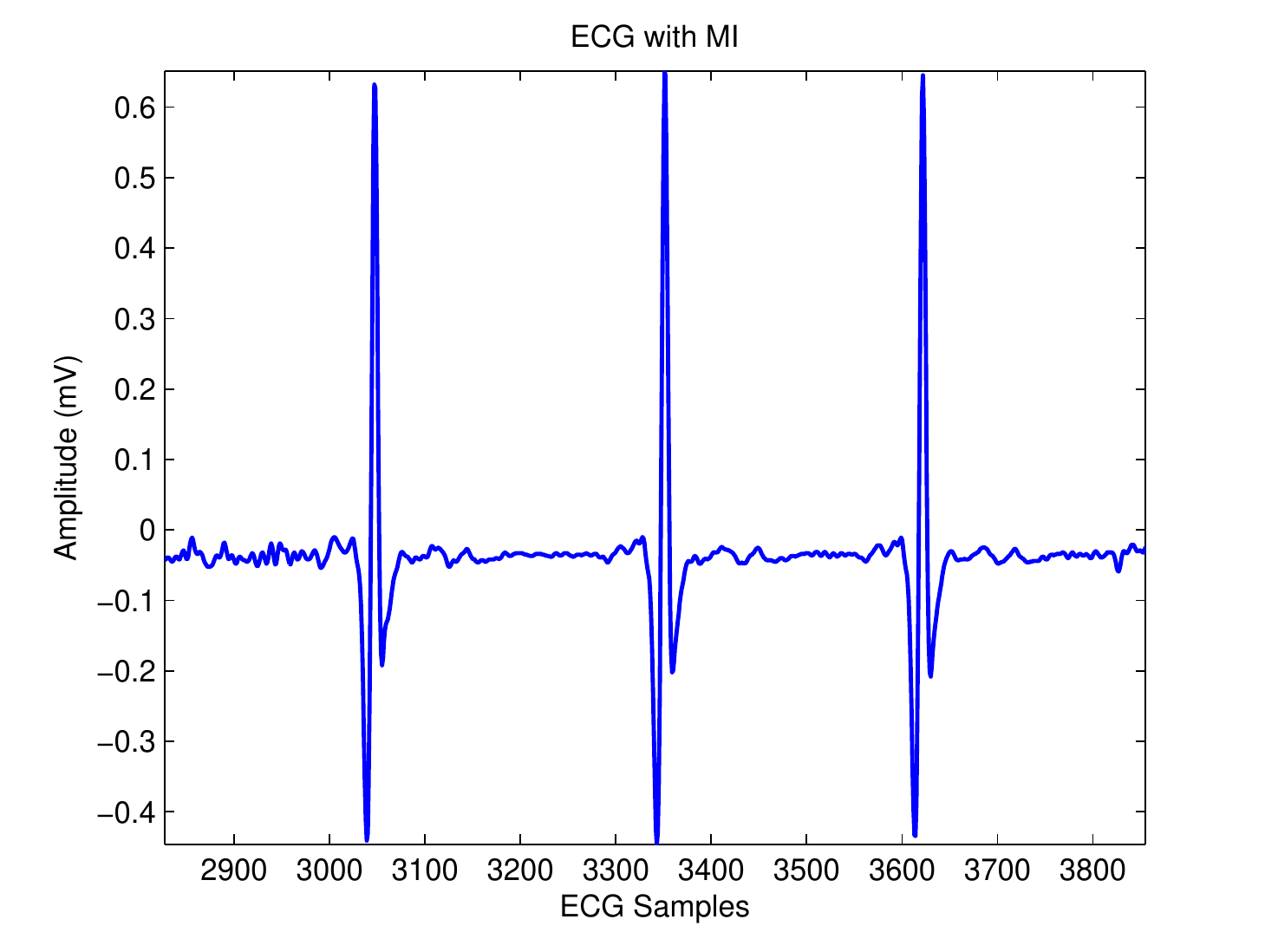}
\label{fig:MI_ecg_processed}}}
\parbox{0.45\textwidth}{
\subfigure[PVC ECG]{\includegraphics[scale=0.43]{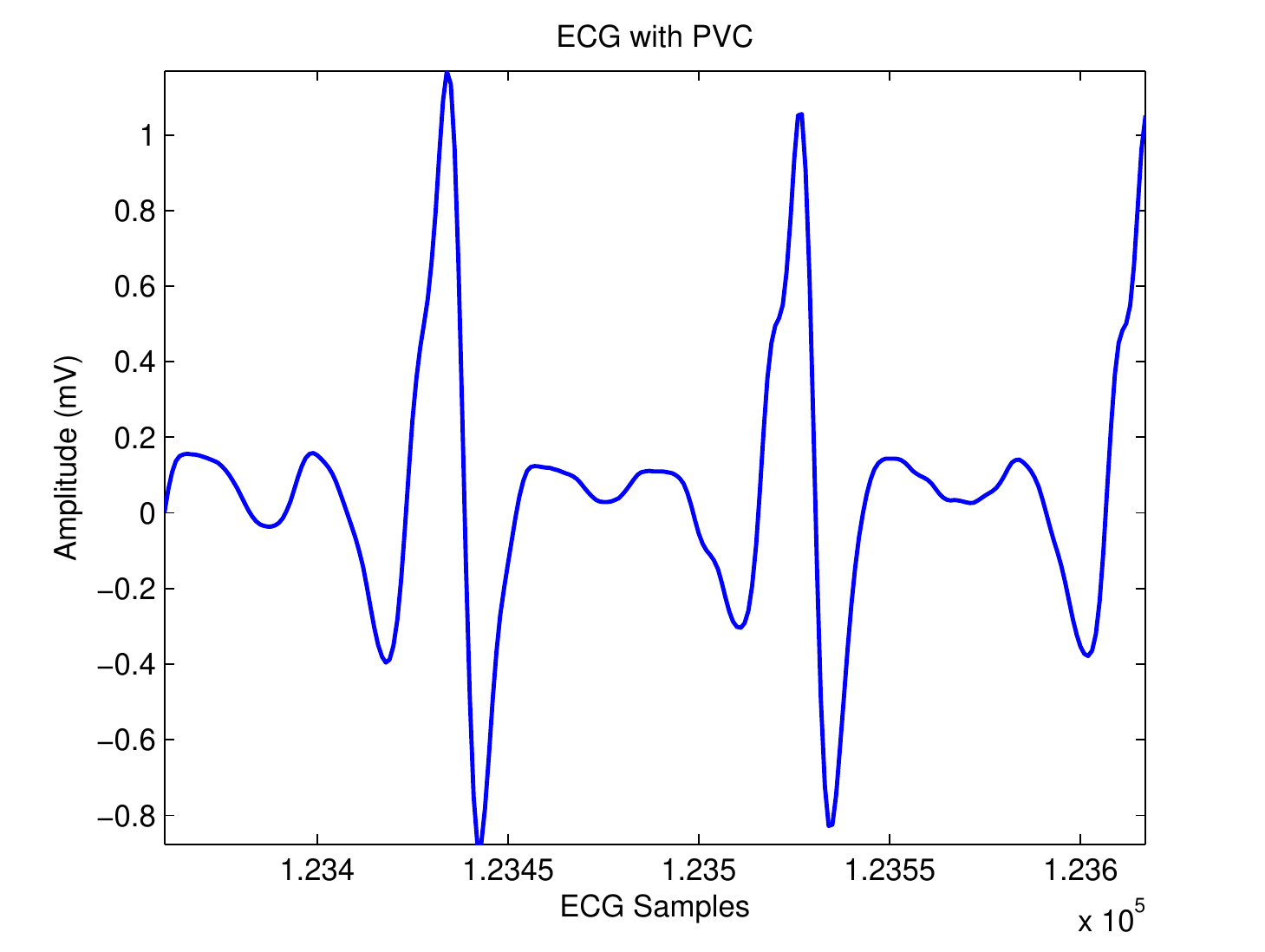}
\label{fig:PVC_ecg_processed}}}
\parbox{0.45\textwidth}{
\subfigure[PAC ECG]{\includegraphics[scale=0.43]{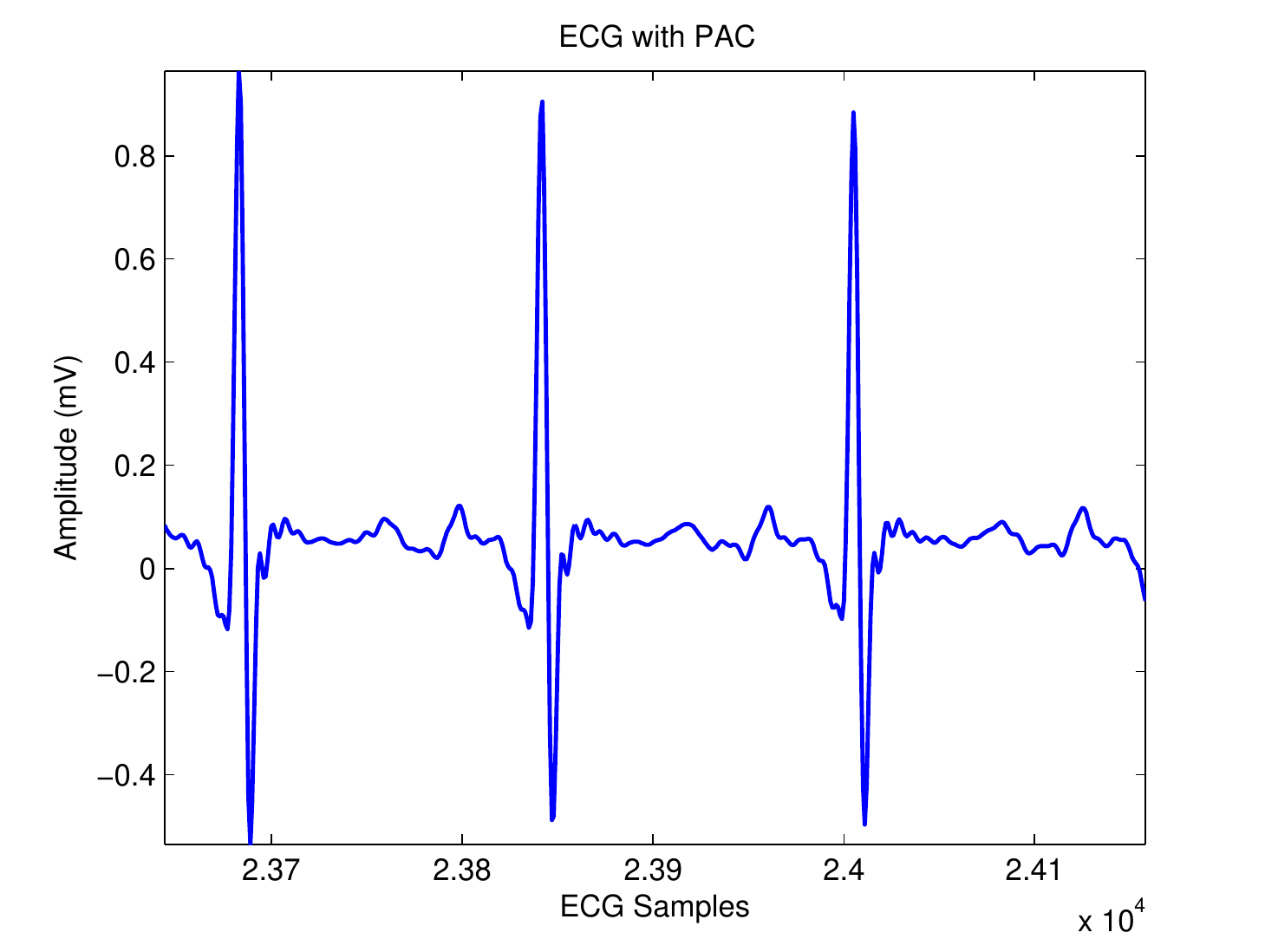}
\label{fig:PAC_ecg_processed}}}
\caption{ECG Pre-processing stage}
\label{fig:processed_ecg}
\end{figure}

Figure~\ref{fig:processed_ecg}  portrays the processed ECG signal by 0.5 Hz, succession of various high pass, low pass filters, and DWT. The goal of preprocessing is to remove the noise from the digitized signal without losing the main characteristics of the signal as depicted in the figure. One can notice in Figure ~\ref{fig:processed_ecg}  that we have the slight rambling baseline in the resultant preprocessed ECG signal without losing the main features of the signal. Moreover, the preprocessing step removes the power line interference,  baseline wandering,  EMG Noise and motion artifacts of electrodes.\par 

%%%%%%%%%%%%%%%%%%%%%%%%FIG PRE PROCESSED %%%%%%%%%%%%%%%%%%%
%%%%%%%%%%%%%%%%%%%%%%%%%%%%
%%%%% Feature Extraction details %%%

\begin{figure}[!tb]  %width=2.5in
\centering
\parbox{0.45\textwidth}{
\subfigure[Normal ECG]{\includegraphics[scale=0.43]{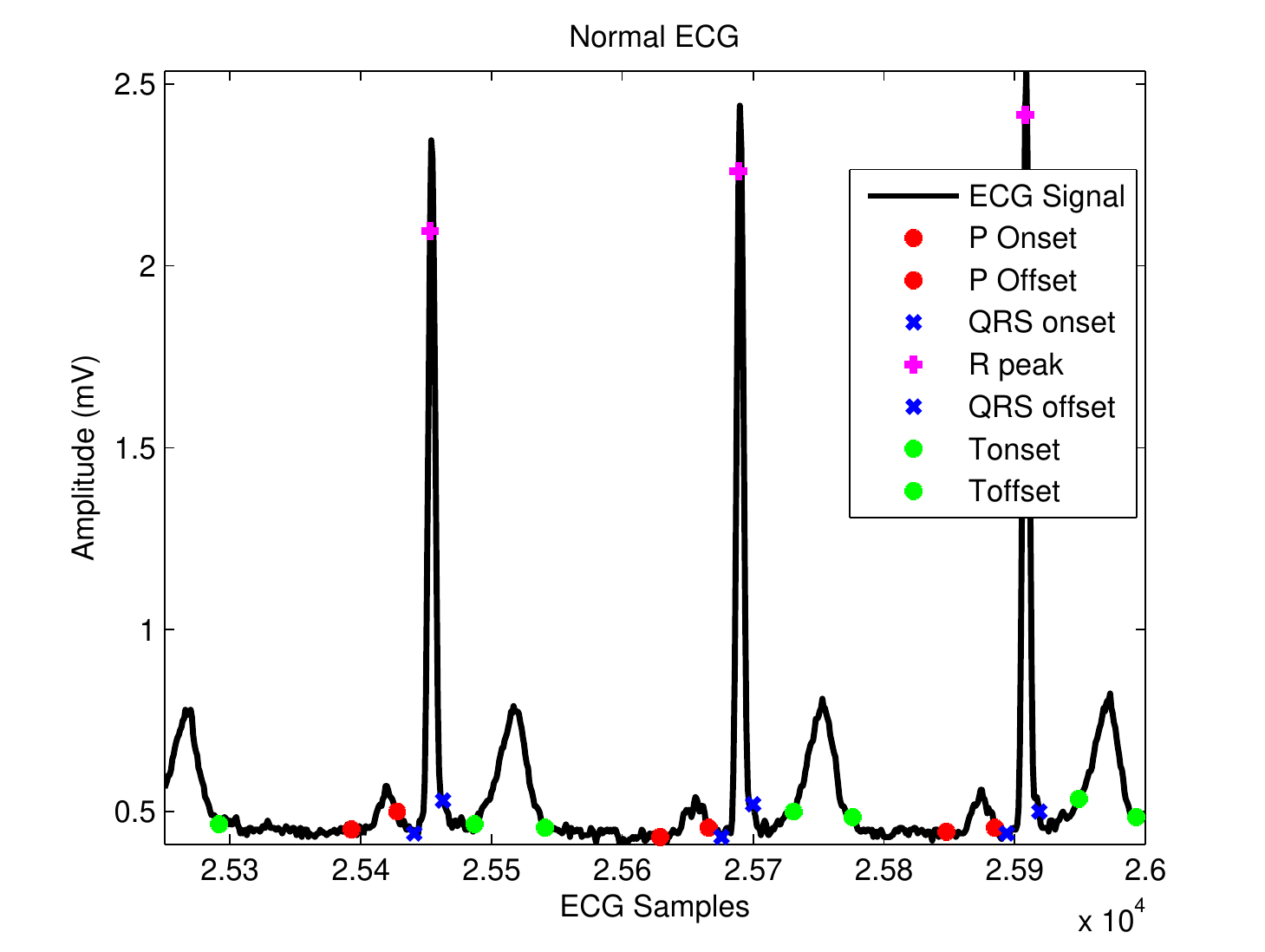}
\label{fig:normal_ecg_features}}}
\parbox{0.45\textwidth}{
\subfigure[MI ECG]{\includegraphics[scale=0.43]{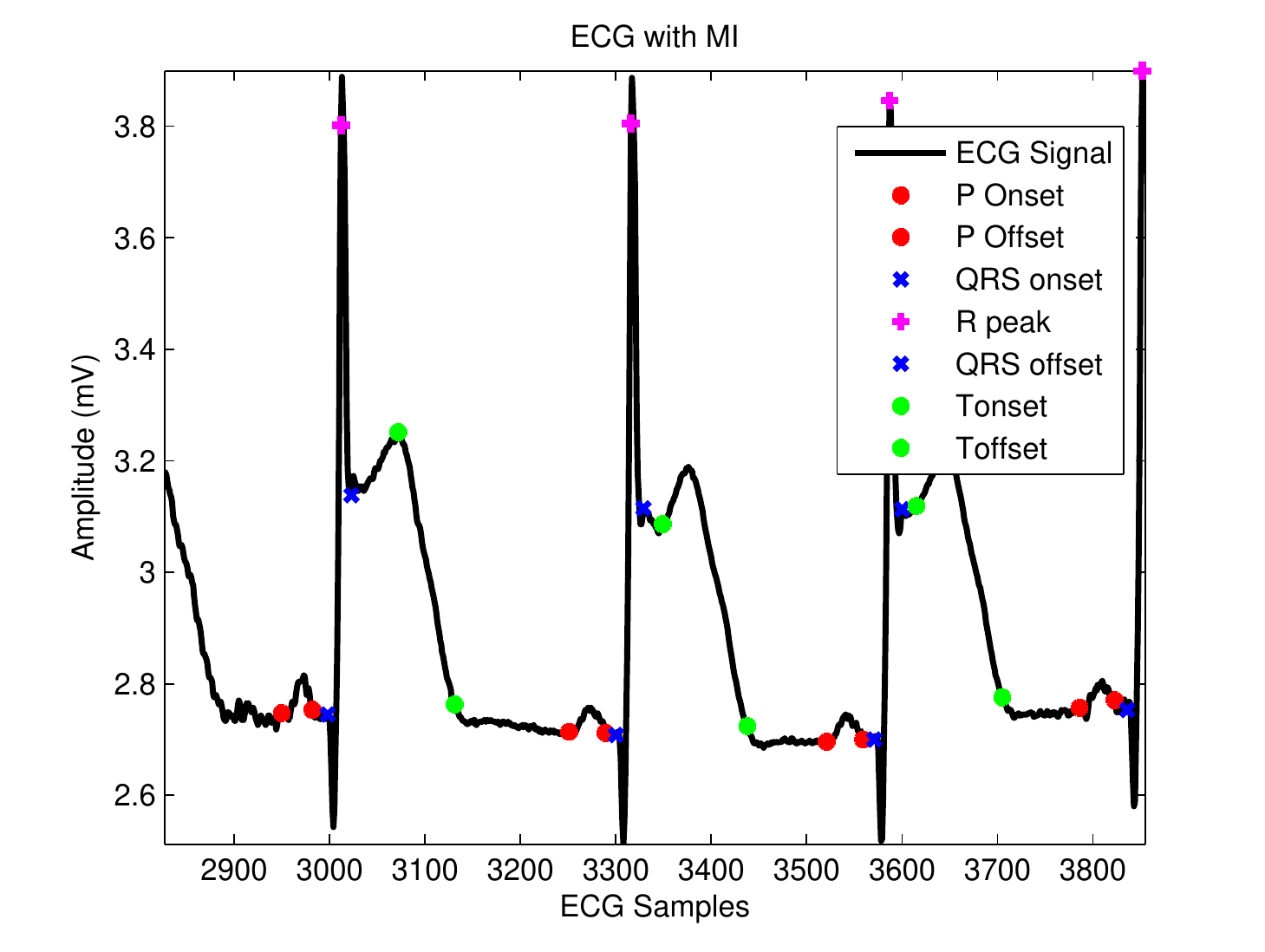}
\label{fig:MI_ecg_features}}}
\caption{ECG features extraction stage}
\label{fig:features_ecg}
\end{figure}

Figure~\ref{fig:features_ecg} depicts the extracted features from the preprocessed $QRS$ complex and then extracted features from $P$ and $T$ peaks of the original wave by applying the UWT as mentioned in Section~\ref{sec:approach}. The extracted $P_{onset}$, $P_{offset}$, $R_{peak}$, $QRS_{onset}$, $QRS_{offset}$, $T_{onset}$ and $T_{offset}$ from two previous ECGs (normal and MI) can be seen in Figure ~\ref{fig:features_ecg}.\par

%%%%%%%%% Extracted Feature FIG %%%%%%%%%%%%%%%%%%%%%%%%%%%
%%%%%%%%%%%%%%%%%%%%%%%%%%%%%%%%%%%%%%%%%%%%%%%%%%%%%%%%%%%%%%

\begin{figure}[!t]
\centering
\parbox{0.45\textwidth}{
\subfigure[Variation of $P_{amp}$]{\includegraphics[scale=0.43]{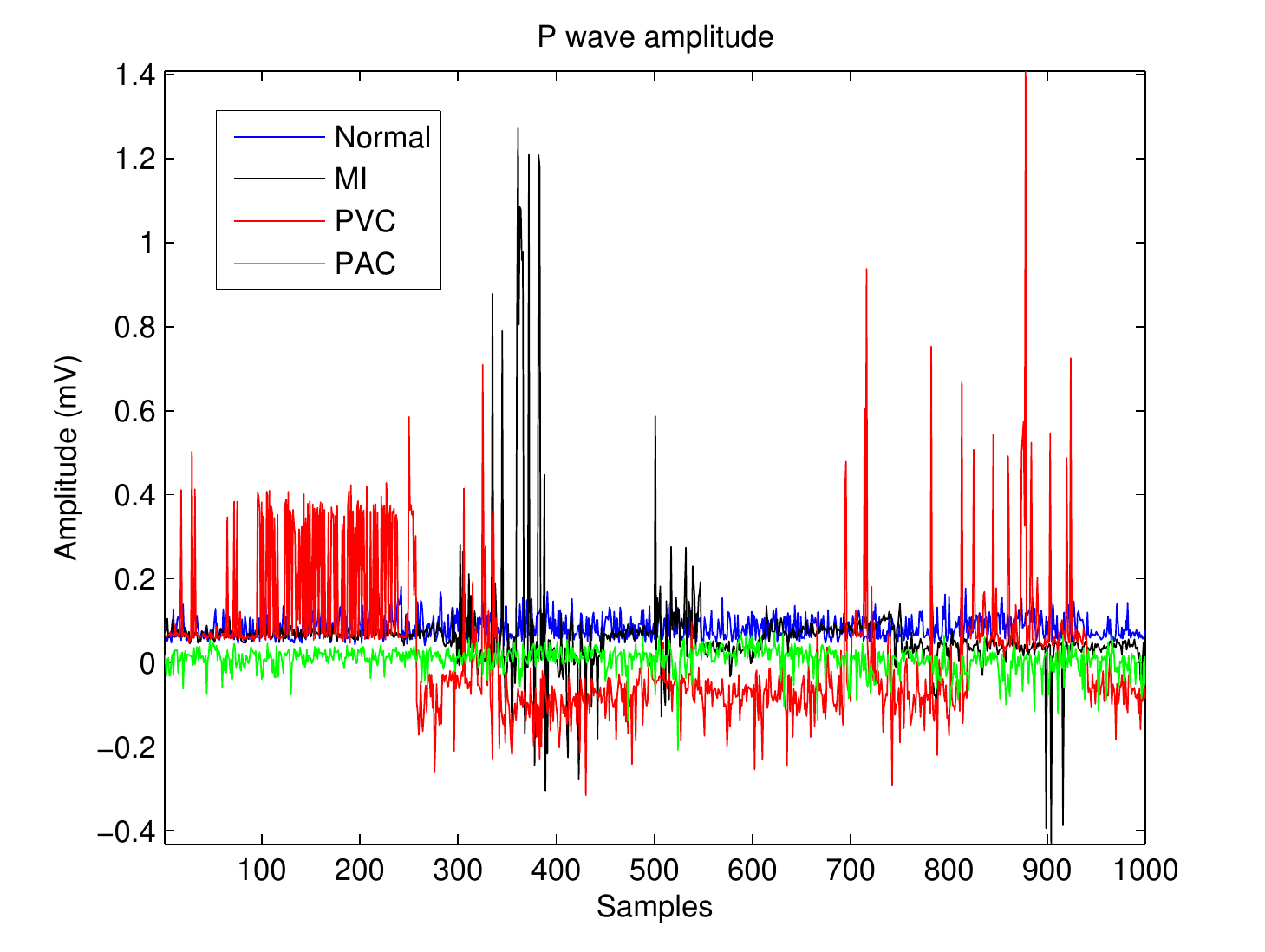}
\label{fig:Pamp_ecg}}
}
\parbox{0.45\textwidth}{
\subfigure[Variation of $QRS_{amp}$]{\includegraphics[scale=0.43]{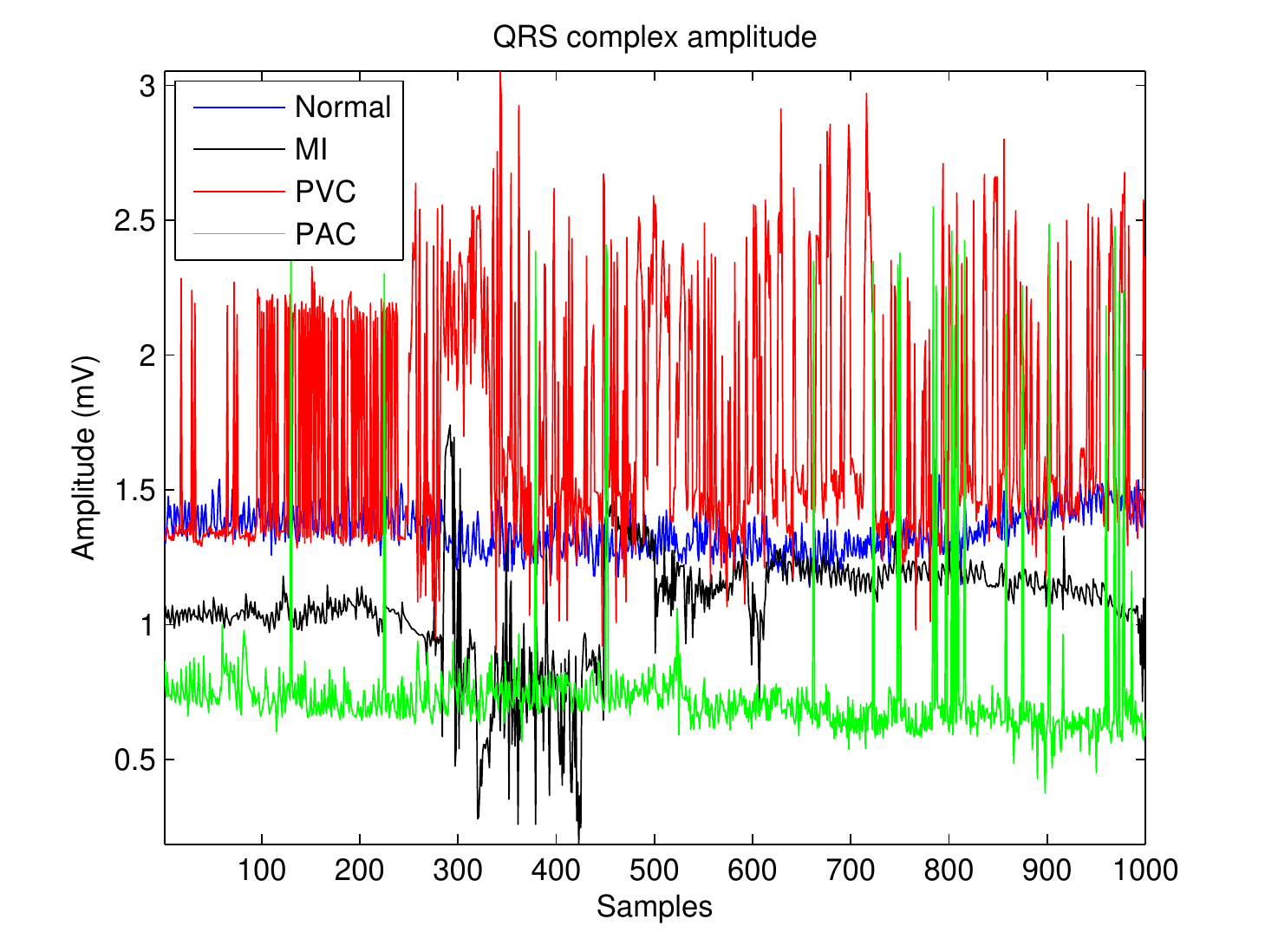}
\label{fig:QRSamp_ecg}}
}
\parbox{0.45\textwidth}{
\subfigure[Variation of $T_{amp}$]{\includegraphics[scale=0.43]{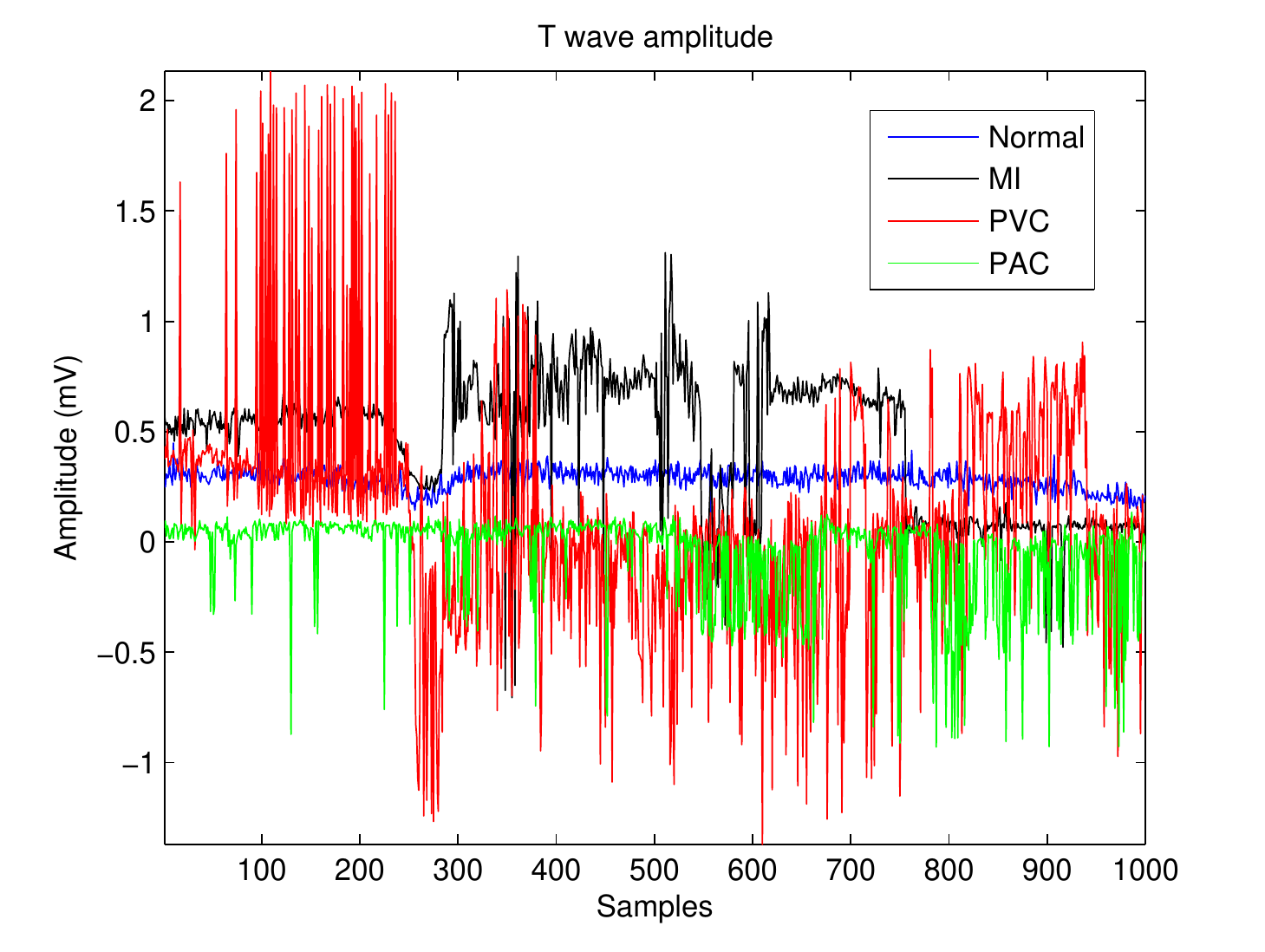}
\label{fig:Tamp_ecg}}
}
\parbox{0.45\textwidth}{
\subfigure[Variation of $QT_{amp}$]{\includegraphics[scale=0.43]{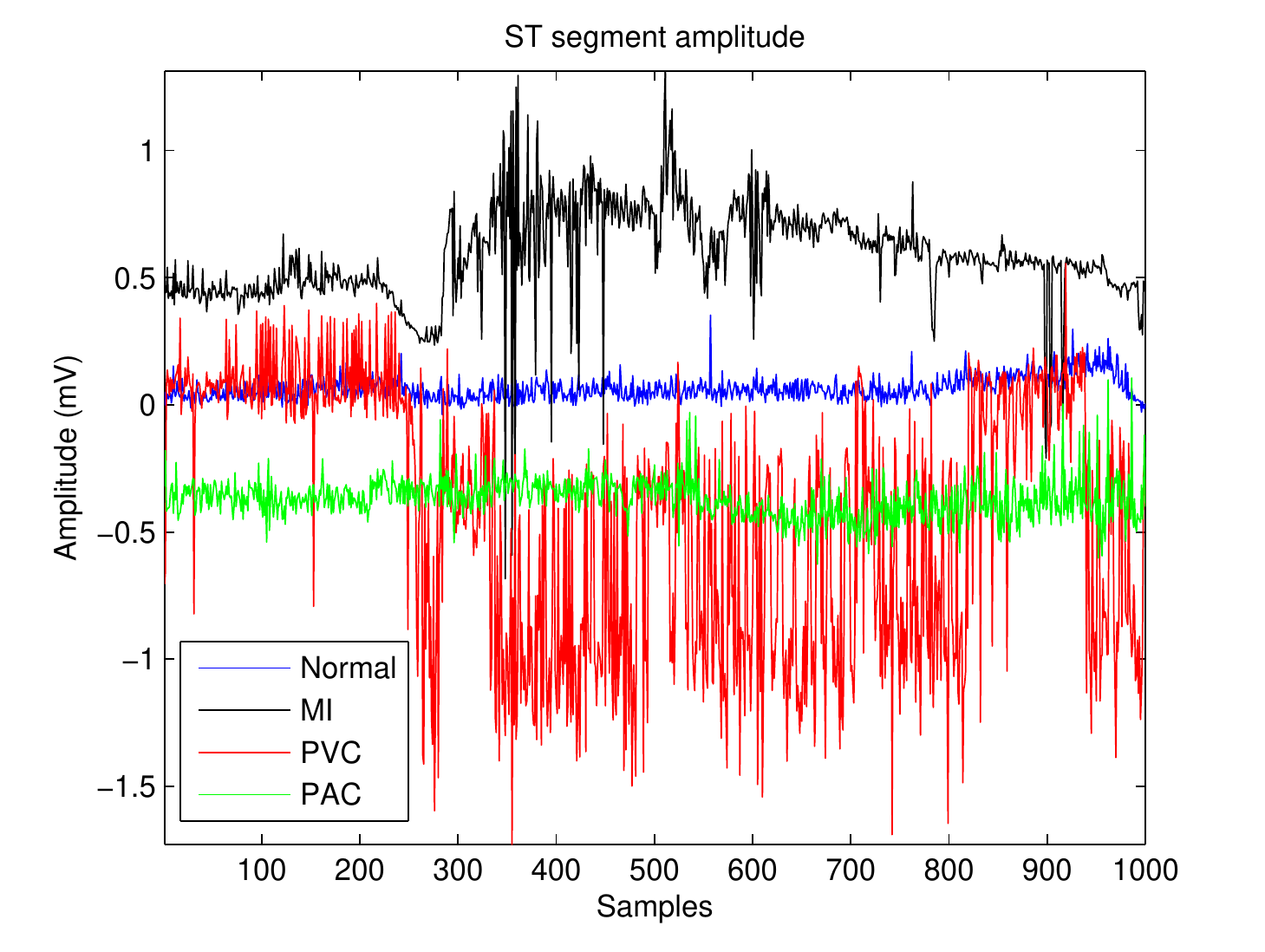}
\label{fig:STamp_ecg}}
}
\caption{ECG amplitude parameters calculation stage}
\label{fig:parameters_amp_ecg}
\end{figure}

%%%%%%%%%%% Amplitude parameter calculation  details%%%%%%%%%%%%%
Figure~\ref{fig:parameters_amp_ecg} depicts the amplitude parameters $P_{amp}$, $QRS_{amp}$, $T_{amp}$, $ST_{amp}$ calculated during the 15 minutes of  ECG recordings in respect of previous Normal, PAC, PVC and MI ECGs. These are ECGs corresponds to 1000 ECG heartbeats. It is obvious in the figure that the amplitude varies significantly for different anomalies while it remains stable for Normal ECG. For instance, ECG for PVC and PAC has the $ST_{amp}$ parameter around $0 mV$ or negative in the figure, while in case of MI the amplitude parameter is greater than $0.5 mV$.\par
%%%%%%%%%%%% Figure Amplitude calculation%%%%%%%%%%%%%%%%%%%%%%%
%%%%%%%%%%%%%%%%%%%%%%%%%%%%%%%%%%%%%%%%%%%%%%%%%%
%%%%%% Details ECG duration parameter stage%%%%%%%

\begin{figure}[!t]
\centering
\parbox{0.45\textwidth}{
\subfigure[Variation of $P_{dur}$]{\includegraphics[scale=0.43]{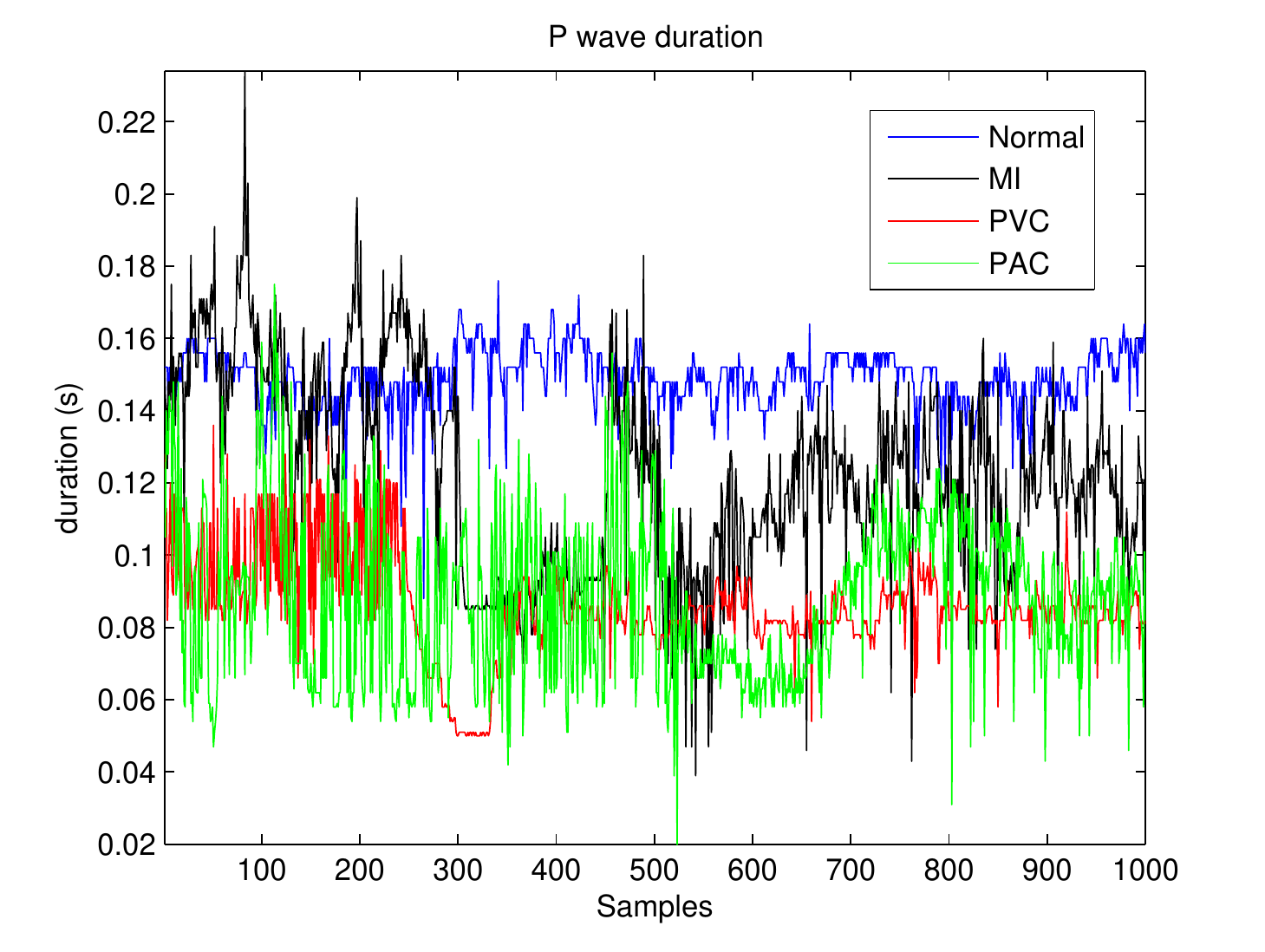}
\label{fig:Pdur_ecg}}}
\parbox{0.45\textwidth}{
\subfigure[Variation of $QRS_{dur}$]{\includegraphics[scale=0.43]{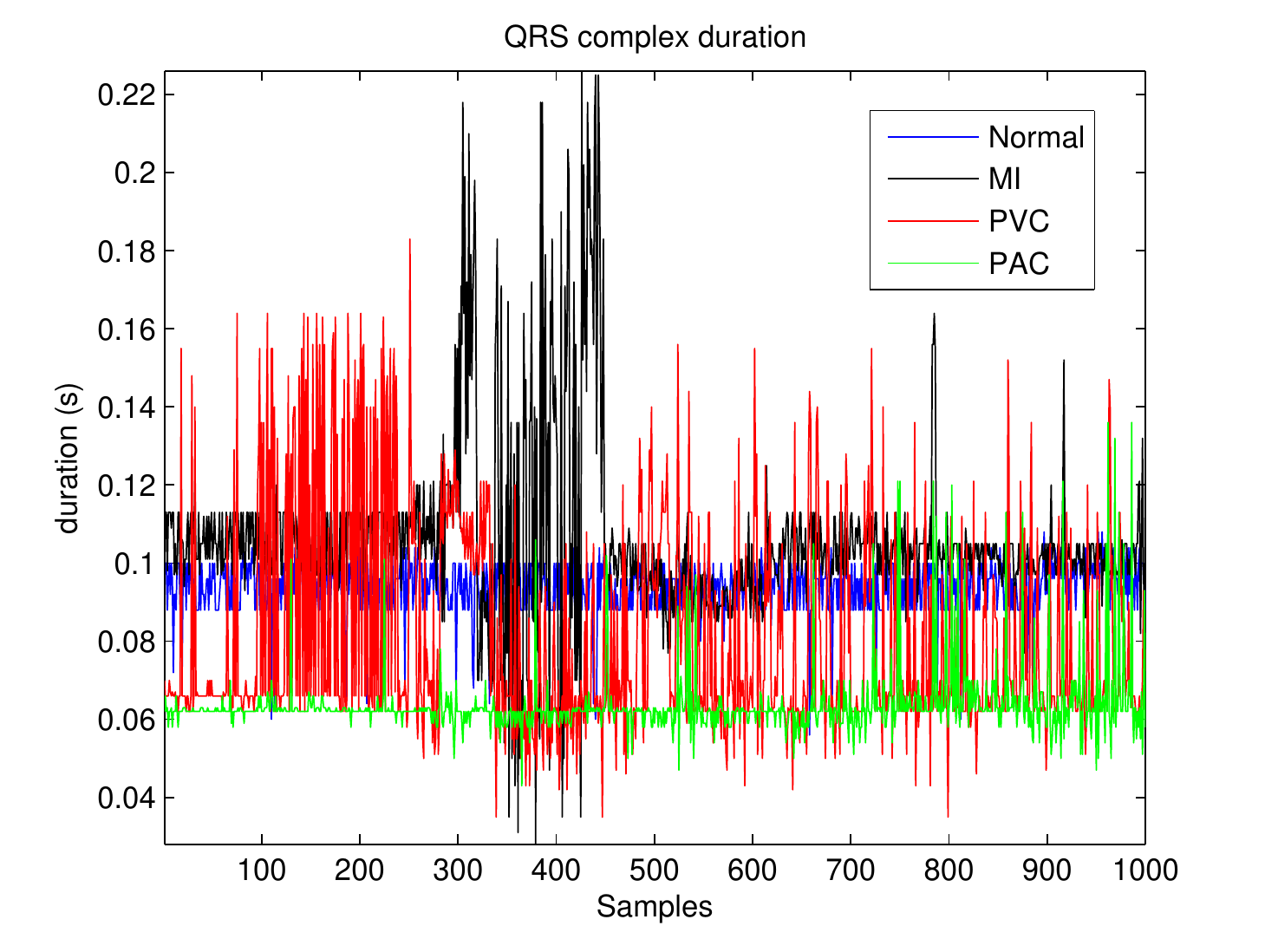}
\label{fig:QRSdur_ecg}}}
\parbox{0.45\textwidth}{
\subfigure[Variation of $T_{dur}$]{\includegraphics[scale=0.43]{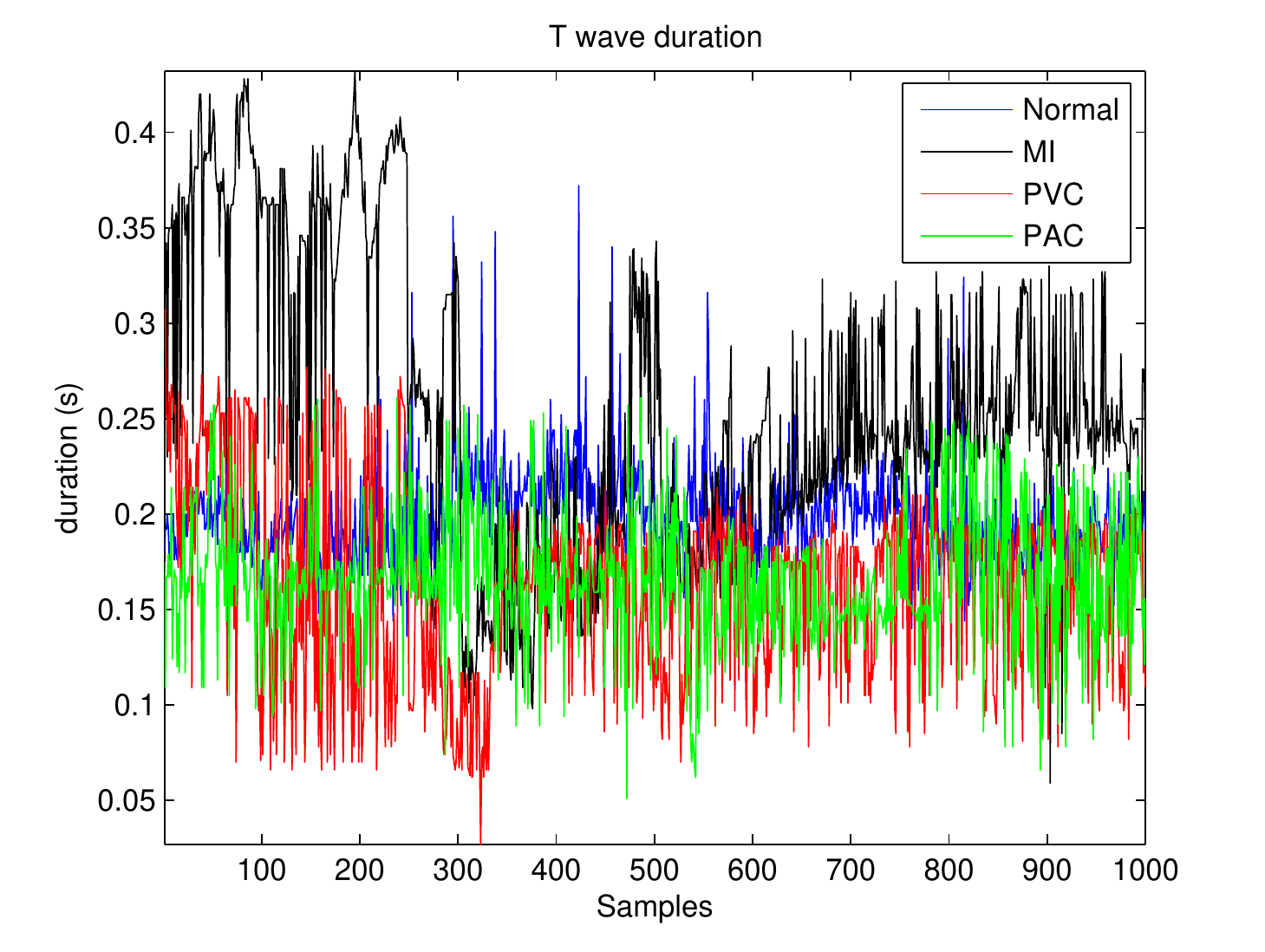}
\label{fig:Tdur_ecg}}}
\parbox{0.45\textwidth}{
\subfigure[Variation of $QT_{dur}$]{\includegraphics[scale=0.43]{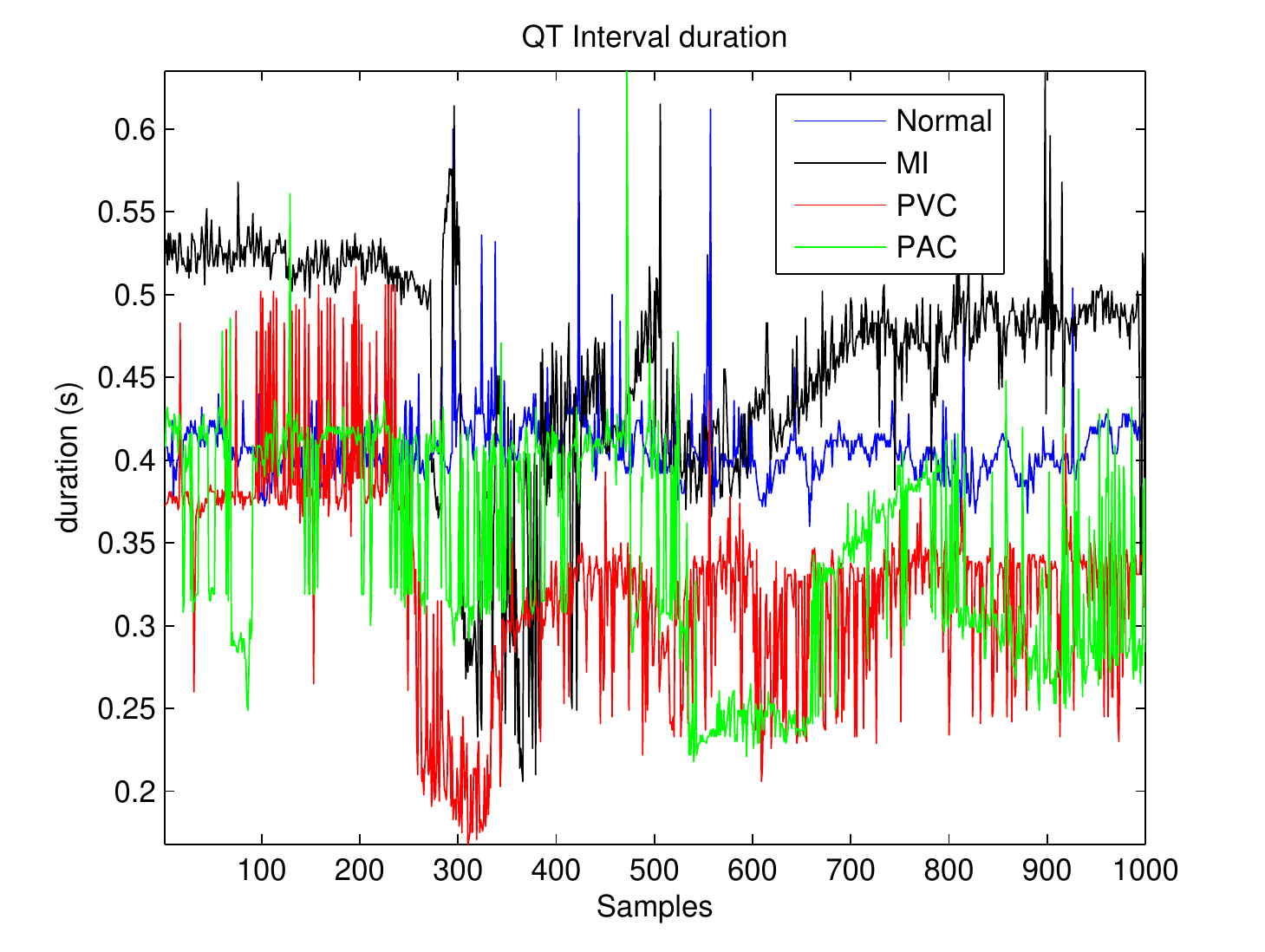}
\label{fig:STdur_ecg}}}
\caption{ECG duration parameters calculation stage}
\label{fig:parameters_dur_ecg}
\end{figure}

Figure~\ref{fig:parameters_dur_ecg} represents the duration parameters $P_{dur}$, $QRS_{dur}$, $T_{dur}$, $PR_{dur}$, $QT_{dur}$ calculated for previous ECGs corresponding to same 1000 ECG heartbeats as discussed above. The difference between Normal and Abnormal ECG is observable in the figure. It can be seen in the figure that the calculated duration parameters, particularly the $QT_{dur}$  and the $T_{dur} $, for ECG with MI abnormality, $QT_{dur}$  is greater than $0.45s$ and the $T_{dur}$ is greater than $0.3s$. Moreover, it can be noticed that these calculated durations for ECG with PAC and PVC are less than $0.4s$ in respect of $QT_{dur}$  and less than $0.25s$ for $T_{dur}$. It is imperative to mention that the amplitude and duration parameters depicted in figures~\ref{fig:parameters_amp_ecg} and~\ref{fig:parameters_dur_ecg} represent the nine statistical variables of our anomaly detection and prediction model using Bayesian Networks for classification and addition of Box plot to remove false alarms.  \par

%%%%%%%%%%%ECG parameter duration calculation Fig%%%%%%%%%%%%%%%
%%%%%%%%%%%%%%%%%%%%%%%%%%%%%%%%%%%%%%%%%%%%%%%%%%%%%%%%%%%%%%%%%%%%%
In order to examine the outcome of our predicted model, we are feeding the preprocessed and nine extracted parameters from all records of two datasets, i.e., EDB and INCARTDB of Physionet database as discussed above.  We have formulated our corpus from these datasets comprised of nine parameters with the anomaly class (PVC, PAC, MI, Normal) of ECG beat. The annotation of anomaly class is provided in the Physionet database against each record. However, further details are available in ~\cite{PhysioAnnot}.  The brief specification of each class in the annotation file is enumerated as under:

\begin{enumerate}[leftmargin=*,labelsep=4.9mm]
\item MI Class: refers to annotated beats ``s'' (ST segment change) or ``T'' (T-wave change). The rationale behind their selection is the medical knowledge~\cite{morris2009abc} according to which  "s" and "T" annotations are the significant symptoms contributing to Myocardial Infarction.
    \item PVC Class: stands for annotated beats ``V''  which refers to a Premature Ventricular Contraction. 
    \item PAC Class: Corresponds to the annotated beats ``A'' refers to Atrial Premature Contraction. 
   \item Normal Class: represents annotated beats ``N'',  which refers to Normal. 
\end{enumerate}
Physionet databases do not necessarily contain all the annotations cited before,  due to this reason we have selected two databases. Furthermore, to obtain more relevant results, we grouped the lead-wise records of each database in order to analyze the performance by lead with a larger number of beats. However, each database contains a different number and type of leads, that can vary from a record to another in the same database.\par 
The EDB dataset contains the ECG records recorded with two leads, but these two leads are different due to which seven leads are distinguishable in the whole database (I, III,  V1,  V2, V3, V4, V5). While in  INCARTDB  dataset posses the record captured by 12 lead ECG. In order to draw fair results comparison, we have selected only above mentioned seven leads from INCARTDB also presents in the EDB dataset. Moreover, to solve the class imbalance problem, we have reduced the proportion of prevailing classes, for instance, Normal Class. Tables~\ref{table:corpus_edb},~\ref{table:corpus_incartdb} depicts the details of the built corpus for each database. \par 
%%%%%%%%%%%%%%%%%%%%%%%%%%%%%%%%
\begin{table} [!htb]
\caption{EDB Corpus}
\centering
\begin{tabular}{|c|c|c|c|c|}
  \hline
  \toprule
   \textbf{ECG Lead} & \textbf{Total records} & \textbf{Total beats} & \textbf{Total ``N'' beats} & \textbf{Total ``MI'' beats} \\  
  \hline
  \midrule
	\textbf{I}  & 19  & 25821 & 12059 & 13762 \\   %&  930
	\hline
	 \textbf{III}& 46  & 69763 & 37597 & 32166 \\   %& 848
	\hline
   \textbf{V1} & 11 & 25302 & 11598 & 13704  \\  %& 273
	\hline
   \textbf{V2} & 10 & 32472 & 15921 & 16551  \\    %& 762
  \hline
	 \textbf{V3} & 7  & 14932 & 8114 & 6818 \\      % & 48
	\hline
	 \textbf{V4} & 34 & 77858 & 38192 & 39666 \\    % & 241
	\hline
	 \textbf{V5} & 51 & 120801 & 61002 & 59799 \\   %& 2254
	\hline
	\bottomrule
	 \end{tabular}
\label{table:corpus_edb}
\end{table}
%%%%%%%%%%%%%%%%%%%%%%%%%%%%%%%%%%%%%%%%%%%%%%%
%%%%%%%%%%%%%%%%%%%%%%%%%%%%%%%
\begin{table} [!htb]
\caption{INCARTDB Corpus}
\centering
%\begin{tabular}{|*{6}{c|}}
\begin{tabular}{|c|c|c|c|c|c|}
  \hline
  \toprule
   \textbf{ECG Lead} & \textbf{Total records} & \textbf{Total beats} & \textbf{Total ``N'' beats} & \textbf{Total ``V'' beats} & \textbf{Total ``A'' beats} \\
	\hline 
	\midrule
	 \textbf{I} & 75 & 37322 & 19008 & 16474 & 1840 \\
	\hline
	 \textbf{III} & 75 & 41380 & 22248 & 17261 & 1871 \\
  \hline
   \textbf{V1} & 75 & 39258 & 20445 & 17064 & 1749 \\
	\hline
   \textbf{V2} & 75 & 38320 & 19877 & 16617 & 1826 \\
  \hline
	 \textbf{V3} & 75 & 40585 & 21695 & 17036 & 1854 \\
  \hline
	 \textbf{V4} & 75 & 40502 & 21228 & 17398 & 1876 \\
	\hline
	 \textbf{V5} & 75 & 41059 & 21769 & 17421 & 1869 \\
	\hline
	\bottomrule
	\end{tabular}
\label{table:corpus_incartdb}
\end{table}
%%%%%%%%%%%%%%%%%%%%%%%%%%%%%%%%%%%%%%%%%%%%%%%%%%%%%%

For the evaluation result, each ECG from the datasets as mentioned earlier is represented by nine extracted parameters. The datasets were split into training and test data. For performance evaluation perspective we are more concerned the performance metrics especially, the overall accuracy of the model, Precision (Positive Prediction Value ), sensitivty (True Positive Rate), Error Rate (Err), False Positive Rate (Far).  The mathematical equations for these metrics (Accuracy, Error rate, sensitivity, Far, Precision ) are given as under:
%%%%%%%%%%%%%%%%%%%%%%%%%%%%%%%%%%%%%%%%%%%%%%%%%%%%%%%%%%%%%%%%%%%%%%%%%
\begin{equation}\label{eq:ACC}
\text{Accuracy(Acc)} =\frac{TP + TN}{TP + TN + FP + FN}
\end{equation}%
\begin{equation}\label{eq:Err}
\text{Error Rate (Err)} =\frac{FP + FN}{TP + TN + FP + FN}
\end{equation}
\begin{equation}\label{eq:TPR}
\text{True Positive Rate (Sensibility)} =\frac{TP}{TP + FN}
\end{equation}
\begin{equation}\label{eq:FAR}
\text{False Positive Rate (Far)} = \frac{FP}{FP + TN}
\end{equation}
\begin{equation}\label{eq:TNR}
\text{True Negative Rate (Specificity)} = \frac{TN}{FP + TN} = {1 - FPR}
\end{equation}
\begin{equation}\label{eq:PPV}
\text{Positive Predictive Value (Precision)} = \frac{TP}{TP + FP}
\end{equation}

 Tables~\ref{table:results_MI},~\ref{table:results_V},~\ref{table:results_A} synthesize the accuracy results obtained for the MI, PVC and PAC classification respectively. These tables are depicting the achieved results comparing to Normal class and three anomalies.  The results comparisons are drawn on seven leads as cited above.\par
\begin{figure}[!htbp]  %width=2.5in
\centering
\parbox{0.45\textwidth}{
\subfigure[ROC Curve for the MI Class]{\includegraphics[scale=0.43]{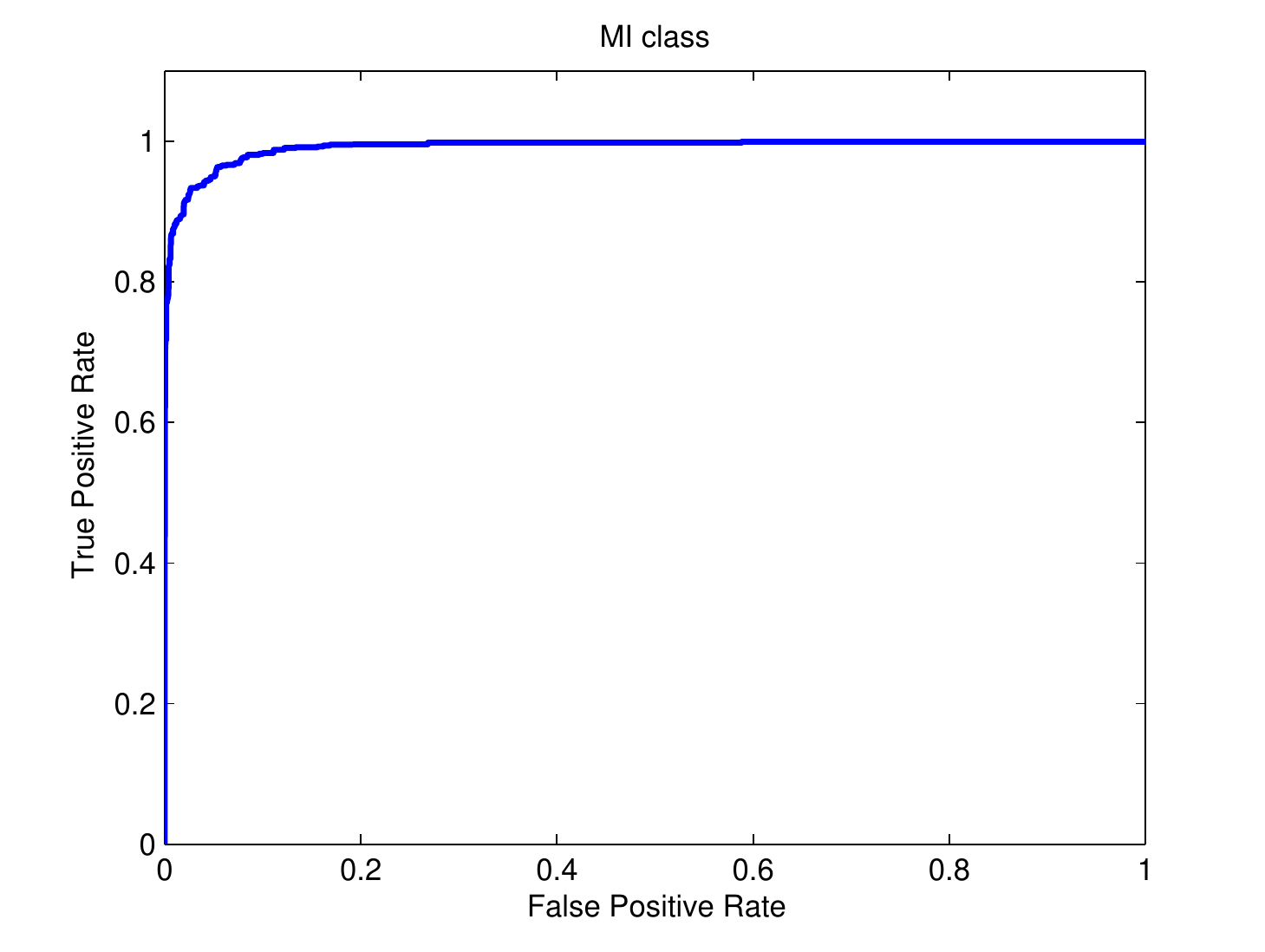}
\label{fig:ROC_MI}}
}
\parbox{0.45\textwidth}{
\subfigure[ROC Curve for the PAC Class]{\includegraphics[scale=0.43]{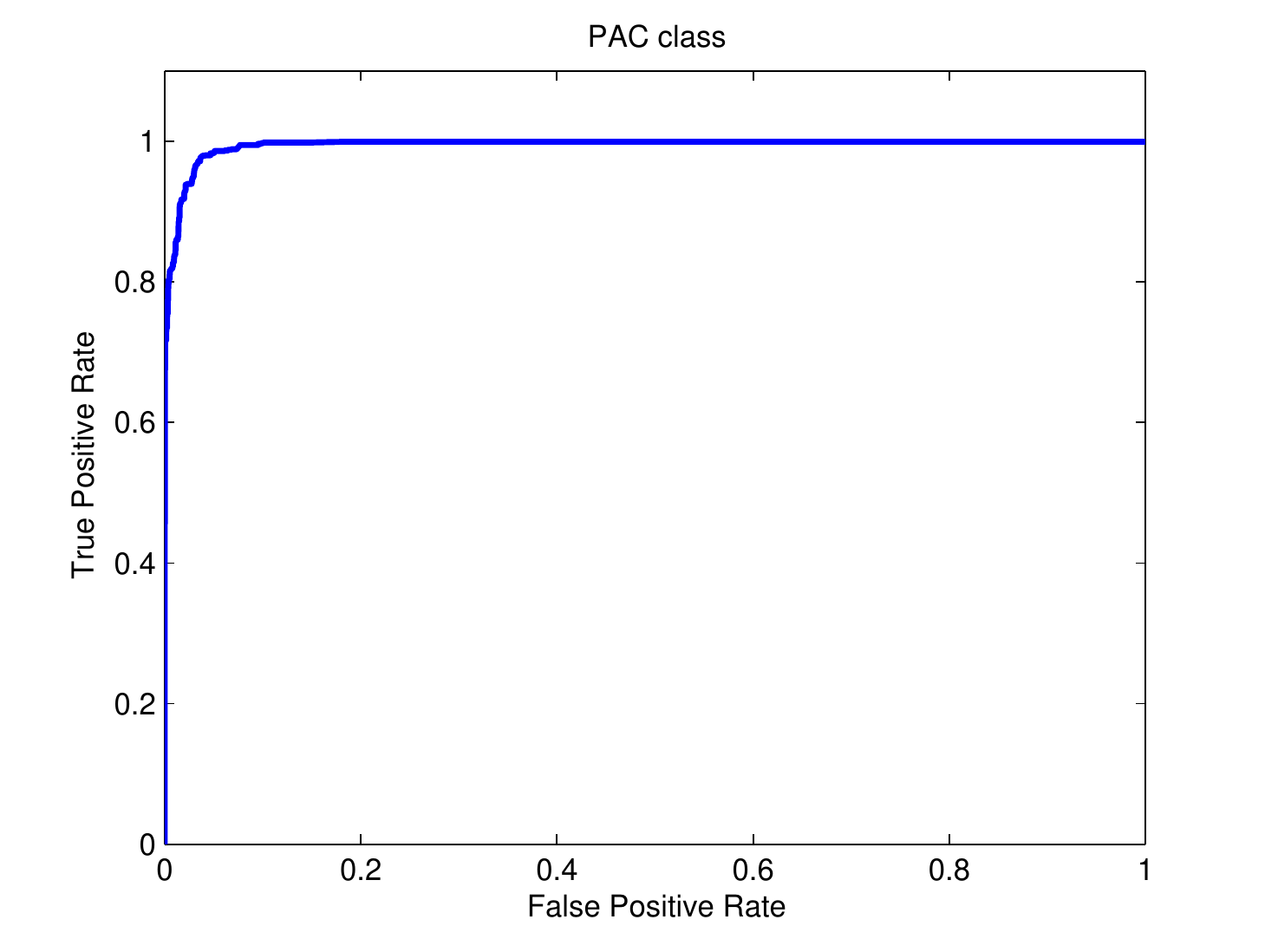}
\label{fig:ROC_PAC}}}
\caption{ROC curves obtained}
\label{fig:roc_ecg}
\end{figure}

%%%%%%%%%%%%%%%%%%%%%%%%%%%%%%%%%%%%%%%%%%%%%%%%%%%%%%%%%%%%%%%%%%%%%%%
Moreover, by using a different size of the training set, we obtain different performance measures for our classification model. For instance,  using  30\% TS, we obtain an average TPR of 98.7\% with 0.5\% FPR. For a TS of 50\%, we obtain an average TPR of 99.3\% with 0.3\% FPR. We periodically update the TS of our system with the new ECG data captured and the results depicted that the achieved accuracy for anomaly classes and lead-wise is phenomenal. Moreover,  minor variation between leads is observable for instance results of lead III are slightly better as compared to others in respect of all four classes.  It is also noticeable that the performance of our prediction model is slightly better for MI and PAC as compared to PVC. The reason behind this lack of performance for PVC is the non-specific variation of ECG parameters as they are more specific in the case of MI and PAC. The same argument holds for Sensibility, Specificity, Error Rate, and False Alarm Rate. Lastly, we have presented the True Positive Rate (TPR) vs. False Positive Rate with different thresholds by capitalizing the Receiver Operating Characteristic (ROC). Figure~\ref{fig:roc_ecg} presents the ROC curve in respect of PAC and MI class.\par

%%%%%%%%%%%%%%%%%%%%%%%%%%%%%%%%%%%5
%%%%%%%%%%%%%%%%%%%%%%%%%%%%%%%%%%%%%

\begin{table} [!htb]
\caption{Prediction model results for MI class}
\centering
\begin{tabular}{|c|c|c|c|c|c|c|}
  \hline
  \toprule
  \textbf{Lead} & \textbf{Acc\%} & \textbf{Err\%} & \textbf{Class} & \textbf{Se\%} & \textbf{Far\%} & \textbf{Prec\%}\\
  \hline
  \midrule
	\hline
	I&93,19\%&6,80\%&MI&93,9\%&7,6\%&93,4\%\\
	\cline{4-7}
	 & & &N&92,4\%&6,1\%&93\%\\
	\hline
	III&94,18\%&5,81\%&MI&93,1\%&4,5\%&96\%\\
	\cline{4-7}
	 & & &N&95,1\%&6,8\%&92,2\%\\   		  
	\hline
	V1&94\%&5,9\%&MI&94,5\%&7,3\%&96,9\%\\
	\cline{4-7}
	 & & &N&92,7\%&5,4\%&87,8\%\\   		
	\hline
	V2&91,7\%&8,3\%&MI&83,3\%&5,6\%&82,7\%\\
	\cline{4-7}
	 & & &N&94,4\%&16,7\%&94,6\%\\   		
	\hline
	V3&92,16\%&7,83\%&MI&93,2\%&9,5\%&94,3\%\\
	\cline{4-7}
	 & & &N&90,4\%&	6,7\%&88,9\%\\    	
	\hline
	V4&92,84\%&7,15\%&MI&92,5\%&6,3\%&96,8\%\\
	\cline{4-7}
	 & & &N& 93,7\%&7,5\%&85,7\%\\   		

	\hline
	V5&92,73\%&7,26\%&MI&92,5\%&6,7\%&96,5\%\\
	\cline{4-7}
	 & & &N&93,3\%&7,5\%&86,3\%\\  		
	\hline
	\bottomrule
\end{tabular}
\label{table:results_MI}
\end{table}

%%%%%%%%%%%%%%%%%%%%%%%%%%%%%%%%%%%%%%%%%%%%%%%
\begin{table} [!htb]
\caption{Prediction model results for PVC class}
\centering
\begin{tabular}{|c|c|c|c|c|c|c|}
  \hline
  \toprule
  \textbf{Lead} & \textbf{Acc\%} & \textbf{Err\%} & \textbf{Class} & \textbf{Se\%} & \textbf{Far\%} & \textbf{Prec\%}\\
  \hline
	\hline
	\midrule
	I&87,32\%&12,6\%&V&83,6\%&9,2\%&89,6\%\\
	\cline{4-7}
	 & & &N&90,8\%&16,4\%&85,4\%\\
	\hline
	III&87,87\%&12,1\%&V&85,7\%&10,2\%&87,9\%\\
	\cline{4-7}
	 & & &N&89,8\%&14,3\%&87,8\%\\   		  
	\hline
	V1&87,62\%&12,3\%&V&84,7\%&9,8\%&88,7\%\\
	\cline{4-7}
	 & & &N&90,2\%&15,3\%&86,8\%\\    		
	\hline
	V2&87,56\%&12,4\%&V&84\%&9,1\%&89,5\%\\
	\cline{4-7}
	 & & &N&90,9\%&16\%&85,9\%\\   		
	\hline
	V3&87,56\%&12,4\%&V&85,6\%&10,7\%&87,4\%\\
	\cline{4-7}
	 & & &N&89,3\%&14,4\%&87,7\%\\
	\hline	
 	V4&85,89\%&14,1\%&V&85,2\%&13,4\%&85,2\%\\
	\cline{4-7}
	 & & &N&86,6\%&14,8\%&86,5\%\\
	\hline		
	V5&86,13\%&13,8\%&V&86,1\%&13,8\%&84,7\%\\
	\cline{4-7}
	 & & &N&86,2\%&13,9\%&87,4\%\\ 		
	\hline
	\bottomrule
\end{tabular}
\label{table:results_V}
\end{table}

%%%%%%%%%%%%%%%%%%%%%%%%%%%%%%%%%%%%%%%%%%%%%%%%%%%%%%%%%
\begin{table} [!htb]
\caption{Prediction model results for PAC class}
\centering
\begin{tabular}{|c|c|c|c|c|c|c|}
  \hline
  \toprule
  \textbf{Lead} & \textbf{Acc\%} & \textbf{Err\%} & \textbf{Class} & \textbf{Se\%} & \textbf{Far\%} & \textbf{Prec\%}\\
  \hline
	\hline
	\midrule
	I&96,54\%&3,45\%&A&96,8\%&3,8\%&96,8\%\\
	\cline{4-7}
	 & & &N&96,2\%&3,2\%&96,2\%\\
	\hline
	III&97,78\%&2,2\%&A&95,6\%&0,5\%&99,3\%\\
	\cline{4-7}
	 & & &N&99,5\%&4,4\%&96,6\%\\   		  
	\hline
	V1&96,53\%&3,4\%&A&97,3\%&4,4\%&96,3\%\\
	\cline{4-7}
	 & & &N&95,6\%&2,7\%&96,9\%\\   		  
	\hline
	V2&95,88\%&4,1\%&A&95,7\%&3,9\%&95,8\%\\
	\cline{4-7}
	 & & &N&96,1\%&4,3\%&95,9\%\\   		  
	\hline
	V3&96,37\%&3,6\%&A&95,7\%&3,1\%&96,5\%\\
	\cline{4-7}
	 & & &N&96,9\%&4,3\%&96,3\%\\   		  
	\hline
	V4&96,68\%&3,3\%&A&95,1\%&1,9\%&97,8\%\\
	\cline{4-7}
	 & & &N&98,1\%&4,9\%&95,7\%\\   		  
	\hline
  V5&96,84\%&3,1\%&A&94\%&0,8\%&99\%\\
	\cline{4-7}
	 & & &N&99,2\%&6\%&95,2\%\\   		  
	\hline
	\bottomrule
\end{tabular}
\label{table:results_A}
\end{table}

%%%%%%%%%%%%%%%%%%%%%%%%%%%%%%%%%%%%%%%%%%%%%%%%%%%%%%%

\section{Conclusion} \label{sec:conclusion}
In this research work, we have proposed a system for the detection and prediction of Cardiac Anomalies.  Our research work is more focused to detect the MI, PAC, PVC anomalies. The proposed system can collect live ECG data from wireless body sensors installed in a WBAN environment. The novel contribution of this research work involves four main parts, i.e., preprocessing of the ECG signal, feature extraction, prediction of cardiac anomaly and verification of anomaly by removing the false alarm. For prerpocessing of the ECG signal, we apply a succession of high and low pass filters, and Discrete Wavelet Transform (DWT) to minimize the signal to noise ratio. For feature extraction, we apply the Undecimated Wavelet Transform (UWT). After preprocessing and feature extraction we apply the Bayesian Belief Network for the prediction Normal and Abnormal beat and then further classification of cardiac anomalies into MI, PAC, and PVC. In addition to BNC, we also apply Tuckey box analysis for the removal of false alarms. For experimental purposes, we have used two annotated datasets from Physionet database, i.e., EDB and INCARTDB of real-time ECGs using seven leads. The experimental results achieved an average accuracy of 96.6\% for PAC, 92.8\% for MI and 87\% for PVC, with an average Error rate of 3.3\% for PAC, 6\% for MI and 12.5\% for PVC. The future work is to prepare the end-to-end prototype using market sensors to validate our experimental results in a user environment.\par

\bibliographystyle{unsrt}  
\bibliography{template}


\end{document}